\documentclass{article}

\usepackage{microtype}
\usepackage{graphicx}
\usepackage{subfigure}
\usepackage{booktabs} 

\usepackage{amsmath,amsfonts,bm}

\def\eqref#1{equation~\ref{#1}}

\def\1{\bm{1}}

\DeclareMathAlphabet{\mathsfit}{\encodingdefault}{\sfdefault}{m}{sl}
\SetMathAlphabet{\mathsfit}{bold}{\encodingdefault}{\sfdefault}{bx}{n}

\newcommand{\E}{\mathbb{E}}

\newcommand{\R}{\mathbb{R}}

\DeclareMathOperator*{\argmax}{arg\,max}

\usepackage{hyperref}

\usepackage{url}

\usepackage[accepted]{icml2020}

\icmltitlerunning{Deep k-NN for Noisy Labels}

\usepackage{algorithm}
\usepackage{algorithmic}

\usepackage{dsfont}
\usepackage{amsmath}
\usepackage{amsfonts}
\usepackage{amssymb}
\usepackage{amsthm}
\usepackage{subfigure}
\usepackage{adjustbox}
\usepackage{epsfig}
\usepackage{color}
\usepackage{enumerate}
\usepackage{placeins}

\newtheorem{theorem}{Theorem}

\newtheorem{definition}{Definition}
\newtheorem{assumption}{Assumption}
\newtheorem{remark}{Remark}

\theoremstyle{definition}

\renewcommand{\P}{\mathbb{P}}

\newcommand{\X}{\mathcal{X}}

\begin{document}

\twocolumn[
\icmltitle{Deep k-NN for Noisy Labels}




\begin{icmlauthorlist}
\icmlauthor{Dara Bahri}{google}
\icmlauthor{Heinrich Jiang}{google}
\icmlauthor{Maya Gupta}{google}
\end{icmlauthorlist}

\icmlaffiliation{google}{Google Research}
\icmlcorrespondingauthor{Dara Bahri}{dbahri@google.com}
\icmlcorrespondingauthor{Heinrich Jiang}{heinrichj@google.com}



\vskip 0.3in
]



\printAffiliationsAndNotice{}  

\begin{abstract}
Modern machine learning models are often trained on examples with noisy labels that hurt performance and are hard to identify. In this paper, we provide an empirical study showing that a simple $k$-nearest neighbor-based filtering approach on the logit layer of a preliminary model can remove mislabeled training data and produce more accurate models than many recently proposed methods. We also provide new statistical guarantees into its efficacy.
\end{abstract}

\section{Introduction}
Machine learned models can only be as good as the data they were used to train on.  With increasingly large modern datasets and automated and indirect labels like clicks, it is becoming ever more important to investigate and provide effective techniques to handle noisy labels.  

We revisit the classical method of filtering out suspicious training examples using $k$-nearest neighbors ($k$-NN) \citep{Wilson:72}.  Like \citet{papernot2018deep}, we apply $k$-NN on the learned intermediate representation of a preliminary model, producing a supervised distance to compute the nearest neighbors and identify suspicious examples for filtering.  In fact, $k$-NN methods have recently been receiving renewed attention for their usefulness \citep{wang2018analyzing,reeve2019fast}. Here, like  \citet{TrustScores}, we use $k$-NN as an auxiliary method to improve modern deep learning.

The main contributions of this paper are:
\begin{itemize}
\item Experimentally showing that $k$-NN executed on an intermediate layer of a preliminary deep model to filter suspiciously-labeled examples works as well or better than state-of-art methods for handling noisy labels, and is robust to the choice of $k$. 
\item Theoretically showing that asymptotically, $k$-NN's predictions will only identify a training example as clean if its label is the Bayes-optimal label. We also provide finite-sample analysis in terms of the margin and how spread out the corrupted labels are (Theorem~\ref{theo:fixed_delta}), rates of convergence for the margin (Theorem~\ref{theo:rates_delta}) and rates under Tsybakov's noise condition (Theorem~\ref{theo:rates_margin}) with all rates matching minimax-optimal rates in the noiseless setting.
\end{itemize}

Our work shows that even though the preliminary neural network is trained with corrupted labels, it still yields intermediate representations that are useful for $k$-NN filtering.  After identifying examples whose labels disagree with their neighbors, one can either automatically remove them and retrain on the remaining, or send to a human operator for further review. This strategy is also be useful in human-in-the-loop systems where one can warn the human annotator that a label is suspicious, and automatically propose new labels based on its nearest neighbors' labels.

In addition to strong empirical performance, deep $k$-NN filtering has a couple of advantages. Firstly, many methods require a clean set of samples whose labels can be trusted. Here we show that the $k$-NN based method is effective both in the presence and absence of a clean set of samples. Second, while $k$-NN  does introduce the hyperparameter $k$, we will show that deep $k$-NN filtering is stable to the choice of $k$: such robustness to hyperparameters is highly desirable as optimal tuning for this problem is often not available in practice (i.e. when no clean validation set is available).

\section{Related Work}
\begin{figure*}[t]
\begin{tabular}{lll}
   \includegraphics[width=0.30\textwidth]{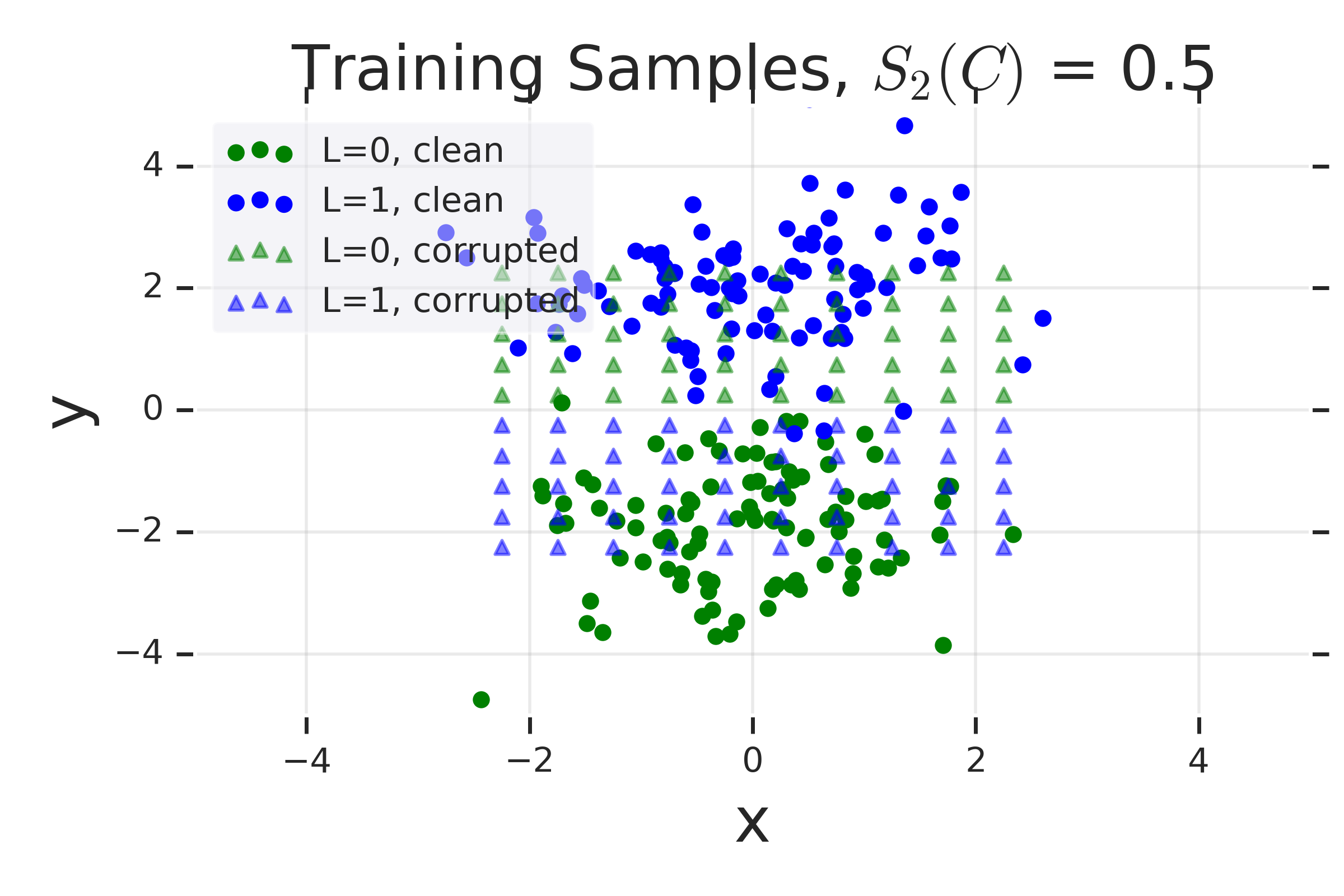} &
   \includegraphics[width=0.30\textwidth]{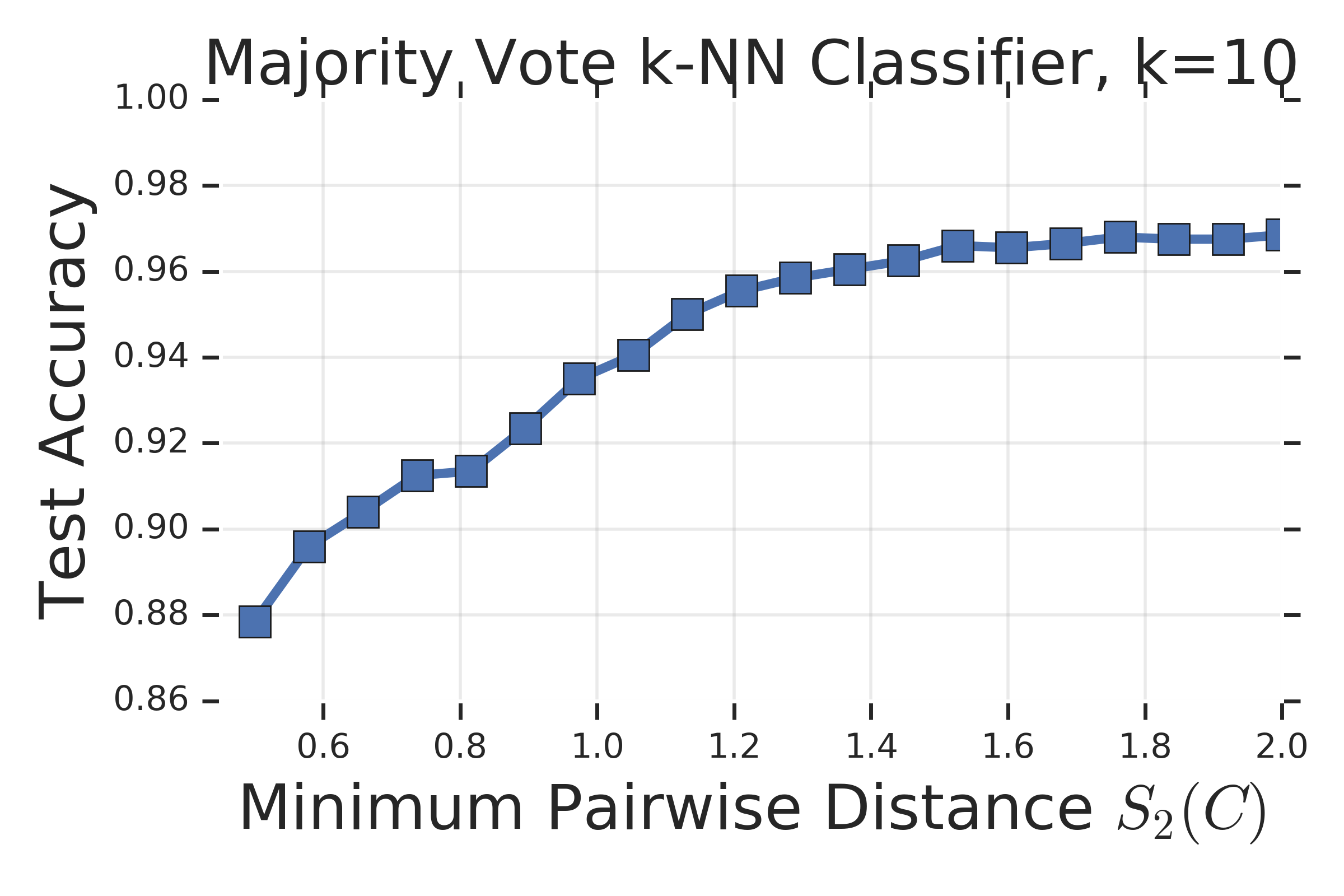} &
   \includegraphics[width=0.30\textwidth]{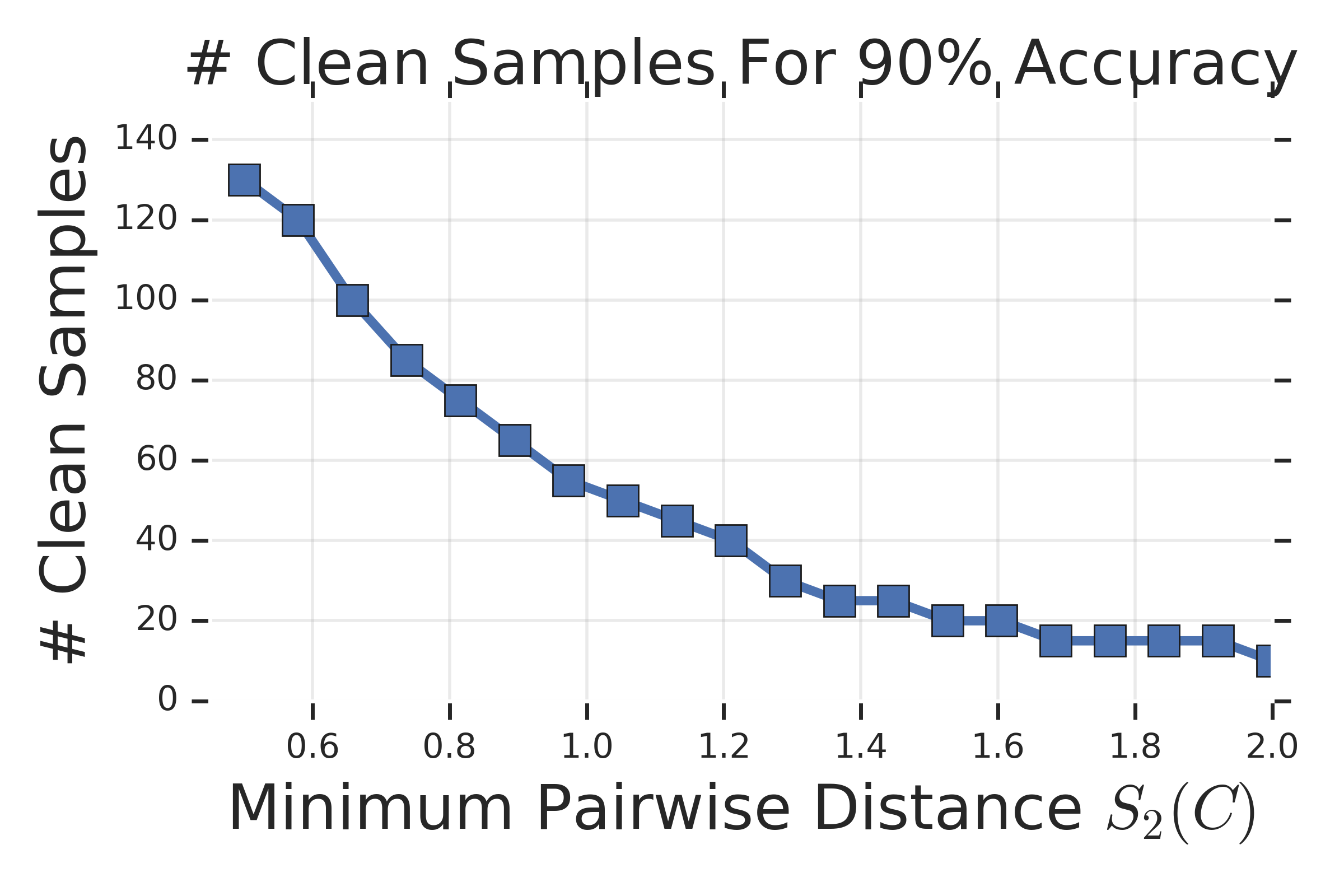}
\end{tabular}
\caption{\label{fig:spread} Left: training samples. We observe that test accuracy improves as $S_2(C)$ increases (middle) and that fewer clean training samples are needed to achieve an accuracy of 90\% (right).}
\end{figure*}

We review relevant prior work for training on noisy labels in Sec. \ref{sec:klang}, and then related $k$-NN theory in \ref{sec:relatedtheory}.

\subsection{Training with noisy labels} \label{sec:klang}
Methods to handle label noise can be classified into two main strategies: (i) explicitly identify and remove the noisy examples, and (ii) indirectly handle the noise with robust training methods. 

{\bf Data Cleaning:} The proposed deep $k$-NN filtering fits into the broad family of \emph{data cleaning} methods, in that our proposal detects and filters \emph{dirty data} \citet{Chu:2016}.  Using $k$-NN to ``edit" training data has been popular since \citet{Wilson:72} used it to throw away training examples that were not consistent with their $k=3$ nearest neighbors.  The idea of using a preliminary model to help identify mislabeled examples dates to at least \citet{Vapnik:1994}, who proposed using the model to compute an information gain for each example and then considered examples with high gain to be suspect. Other early work used a cross-validation set-up to train a classifier on part of the data, then use it to make a prediction on held-out training examples, and remove any examples if the prediction disagrees with the label \citep{Brodley:99}.  

{\bf Noise Corruption Estimation:}
For multi-class problems, a popular approach is to account for noisy labels by applying a confusion matrix after the model's softmax layer \citep{sukhbaatar2014training}.
Such methods rely on a confusion matrix which is often unknown and must be estimated. \citet{patrini2017making} suggest deriving it from the softmax distribution of the model trained on noisy data, but there are other alternatives \citep{goldberger2016training,jindal2016learning,han2018masking}. Accurate estimates are generally hard to attain when only untrusted data is available. \citet{hendrycks2018using} achieves more accurate estimates in the setting where some amount of known clean, trusted data is available. EM-style algorithms have also been proposed to estimate the clean label distribution \citep{xiao2015learning,khetan2017learning,vahdat2017toward}.

{\bf Noise-Robust Training:} \citet{natarajan2013learning} propose a method to make any surrogate loss function noise-robust given knowledge of the corruption rates. \citet{ghosh2017robust} prove that losses like mean absolute error (MAE) are inherently robust under symmetric or uniform label noise while \citet{zhang2018generalized} show that training with MAE results in poor convergence and accuracy. They propose a new loss function based on the negative Box-Cox transformation that trades off the noise-robustness of MAE with the training efficiency of cross-entropy. Lastly, the ramp, unhinged, and savage losses have been proposed and theoretically justified to be noise-robust for support vector machines \citep{brooks2011support,van2015learning,masnadi2009design}. \citet{amid2019robust} construct a noise-robust ``bi-tempered'' loss by introducing temperature in the exponential and log functions. \citet{rolnick2017deep} empirically show that deep learning models are robust to noise when there are enough correctly labeled examples and when the model capacity and training batch size are sufficiently large. \citet{pmlr-v97-thulasidasan19a} propose a new loss that enables the model to abstain from confusing examples during training.

\begin{figure*}[t]
\begin{tabular}{lll}
  \includegraphics[width=0.30\textwidth]{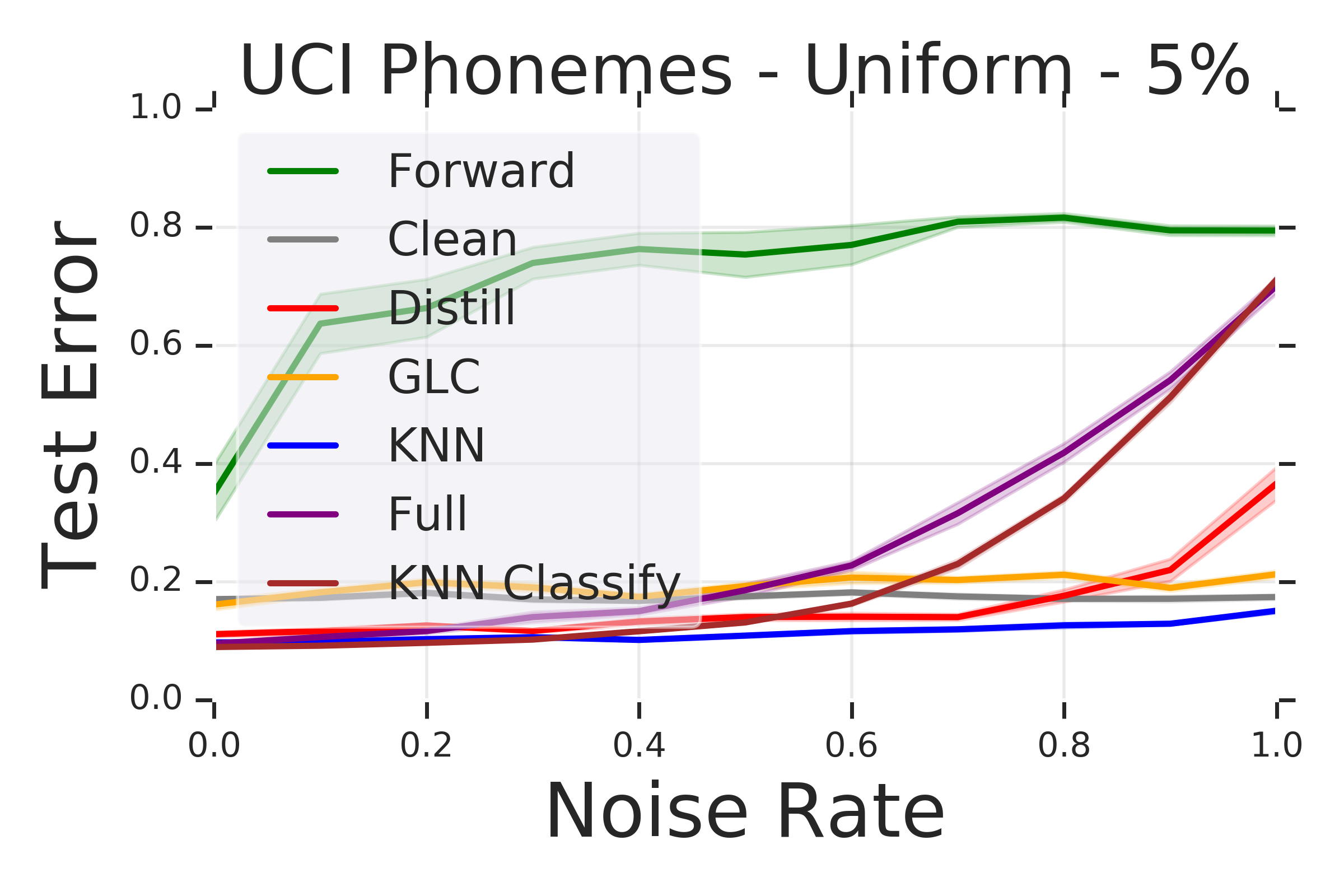} &   \includegraphics[width=0.30\textwidth]{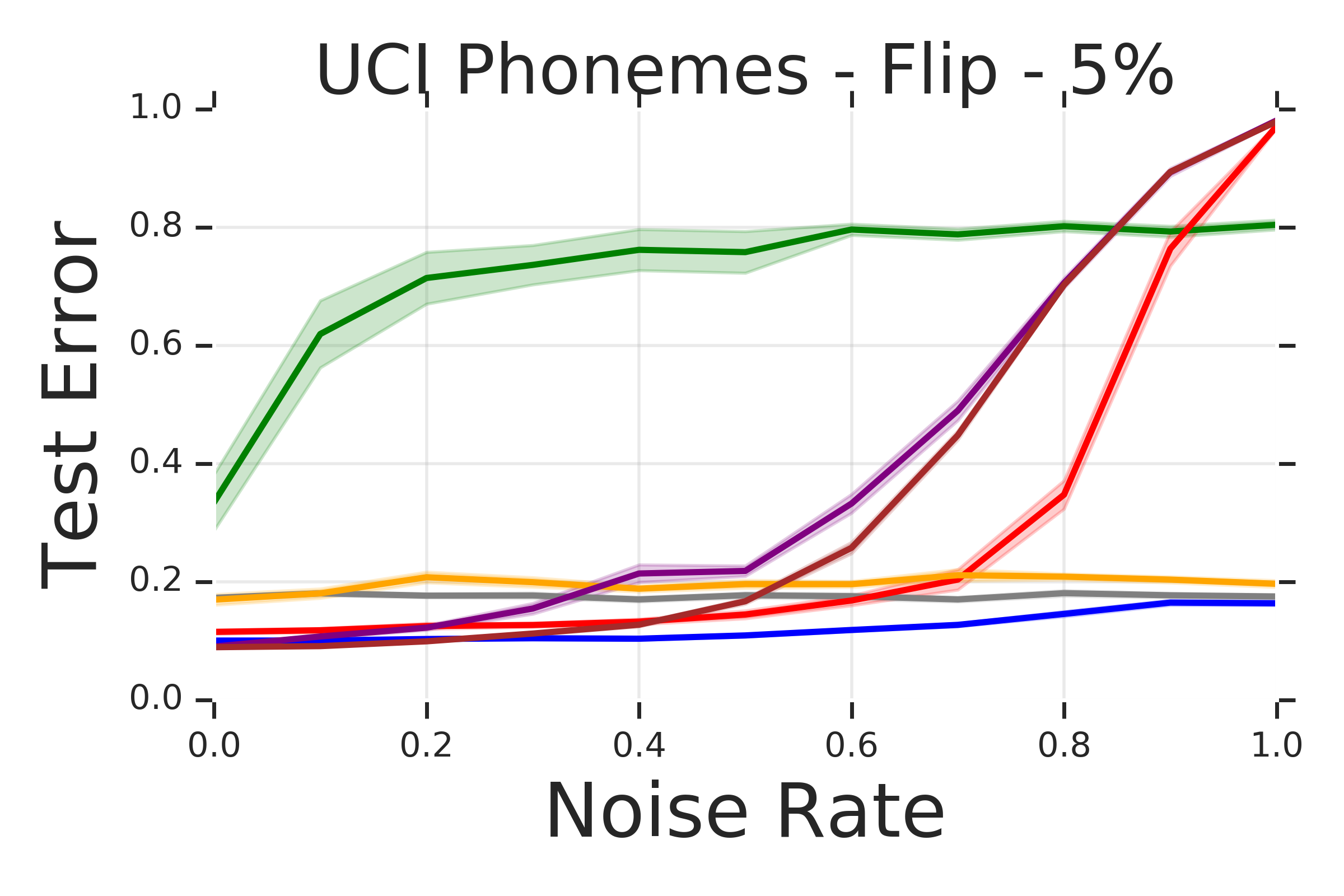}  \includegraphics[width=0.30\textwidth]{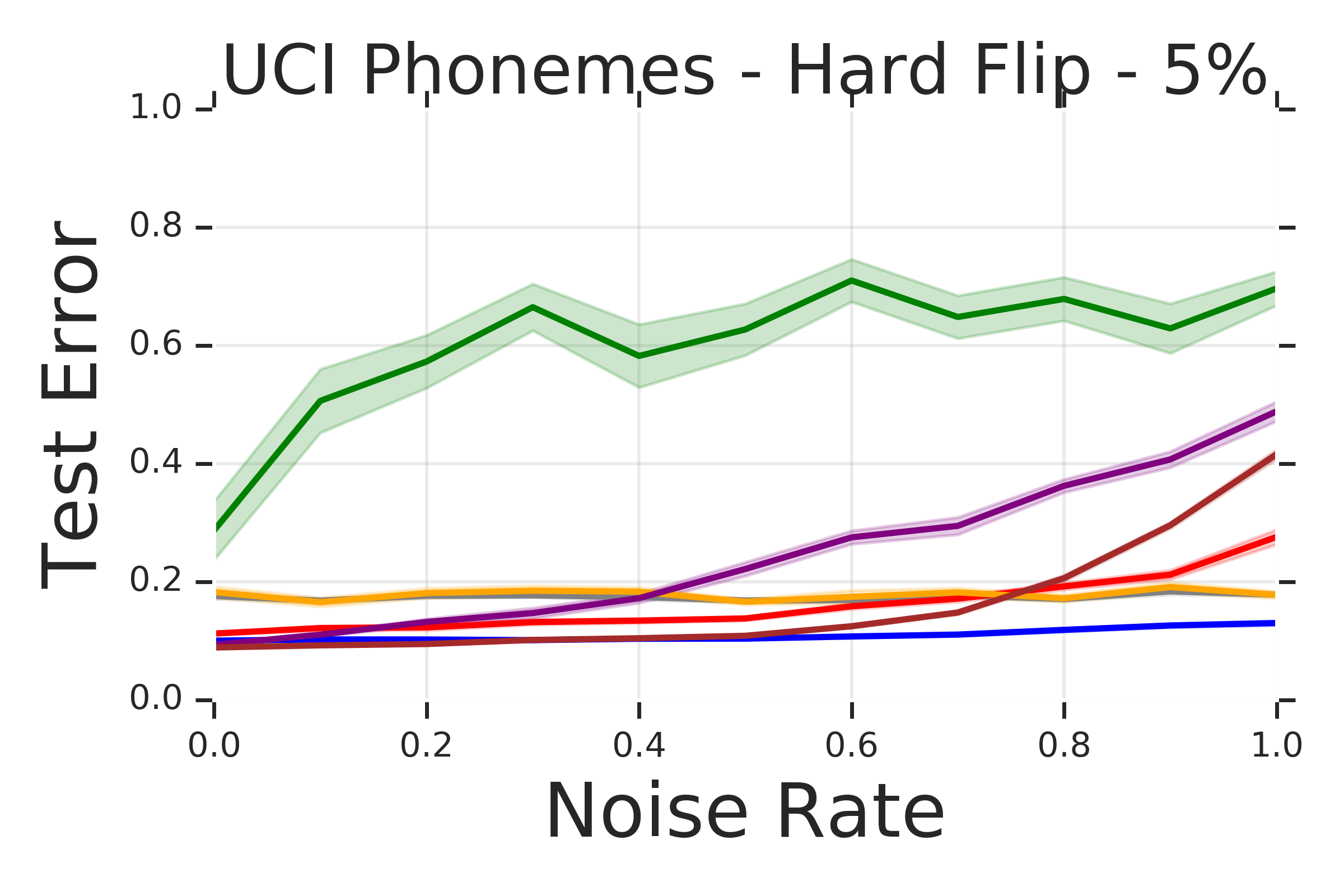} \\
  \includegraphics[width=0.30\textwidth]{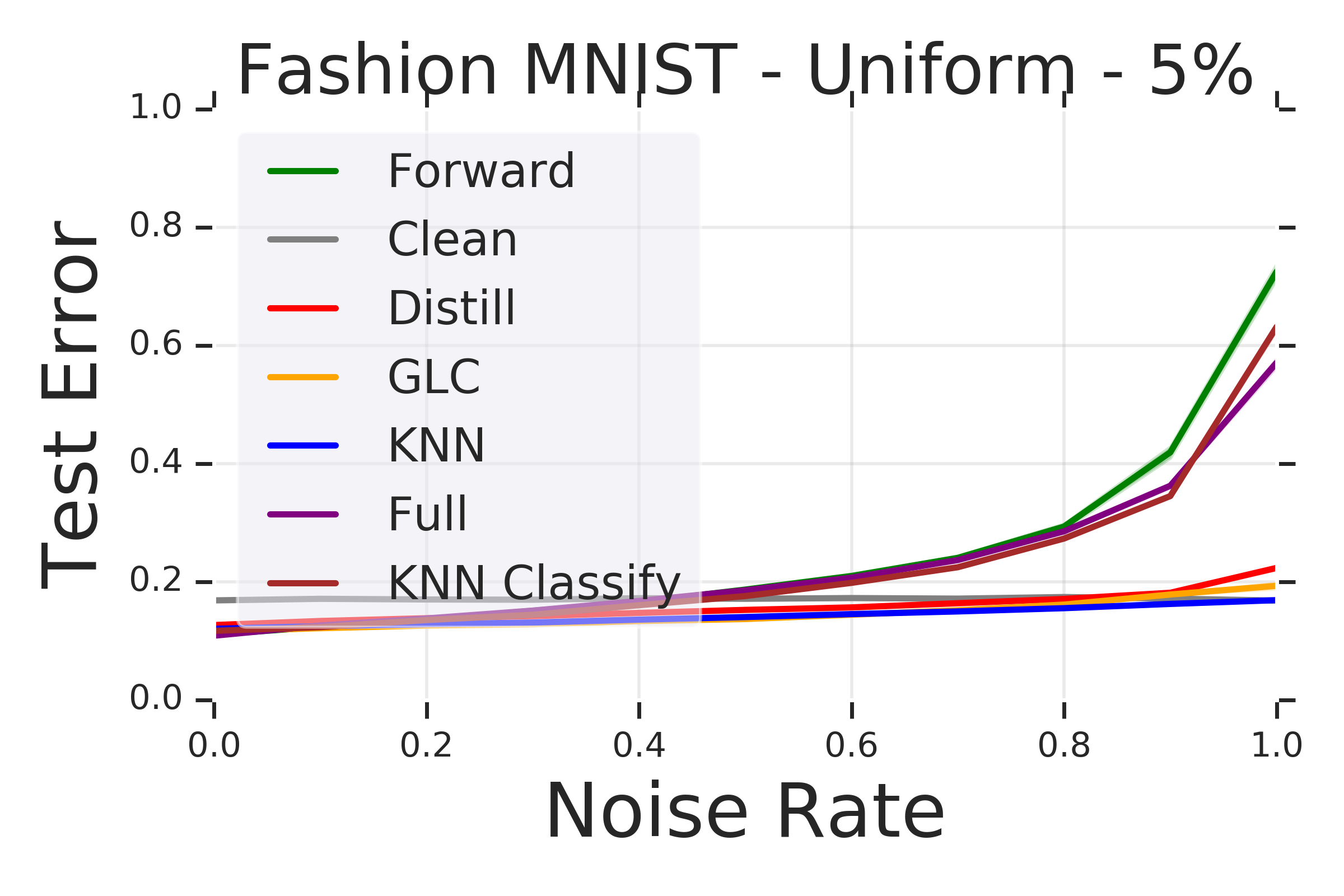} &   \includegraphics[width=0.30\textwidth]{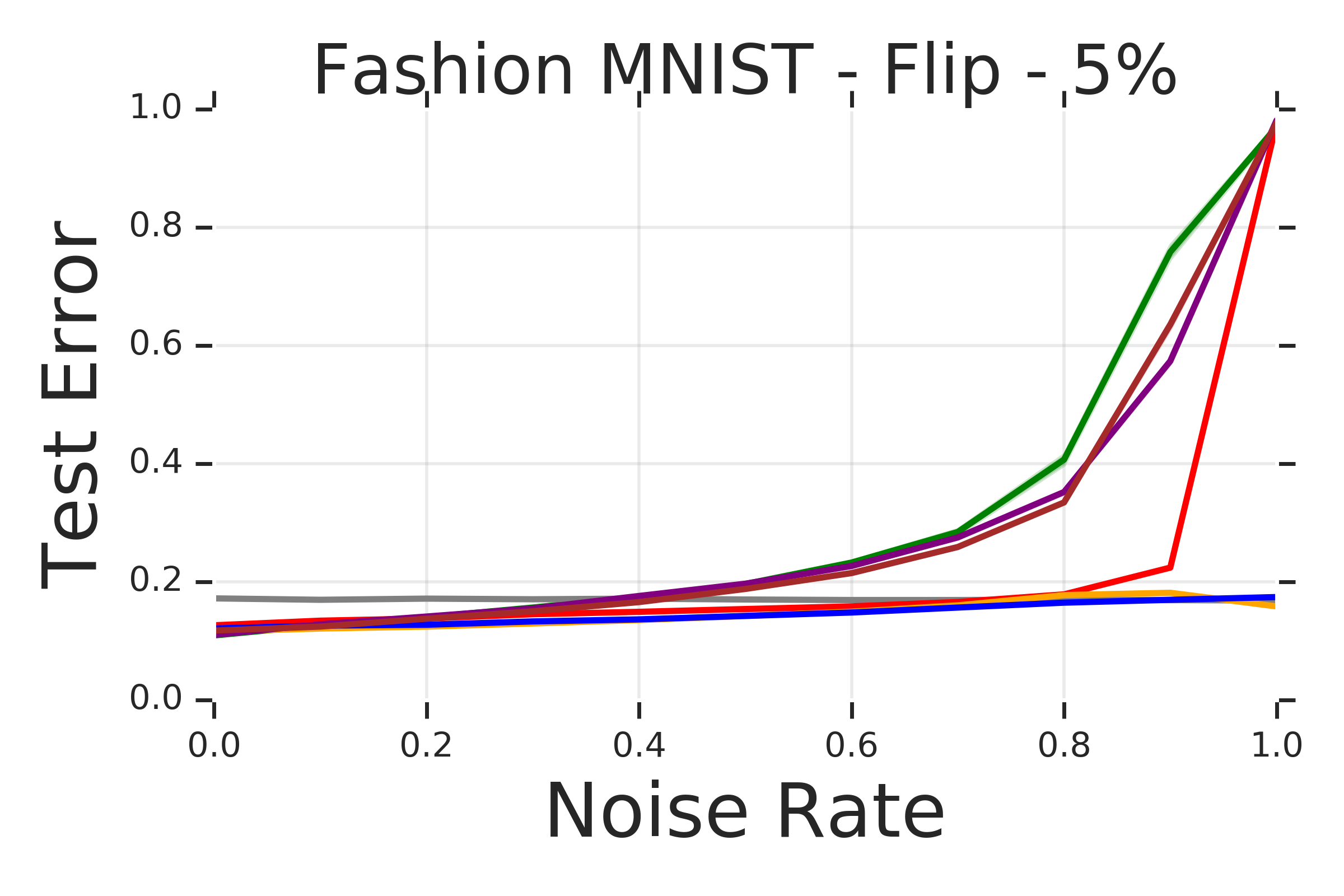} 
 \includegraphics[width=0.30\textwidth]{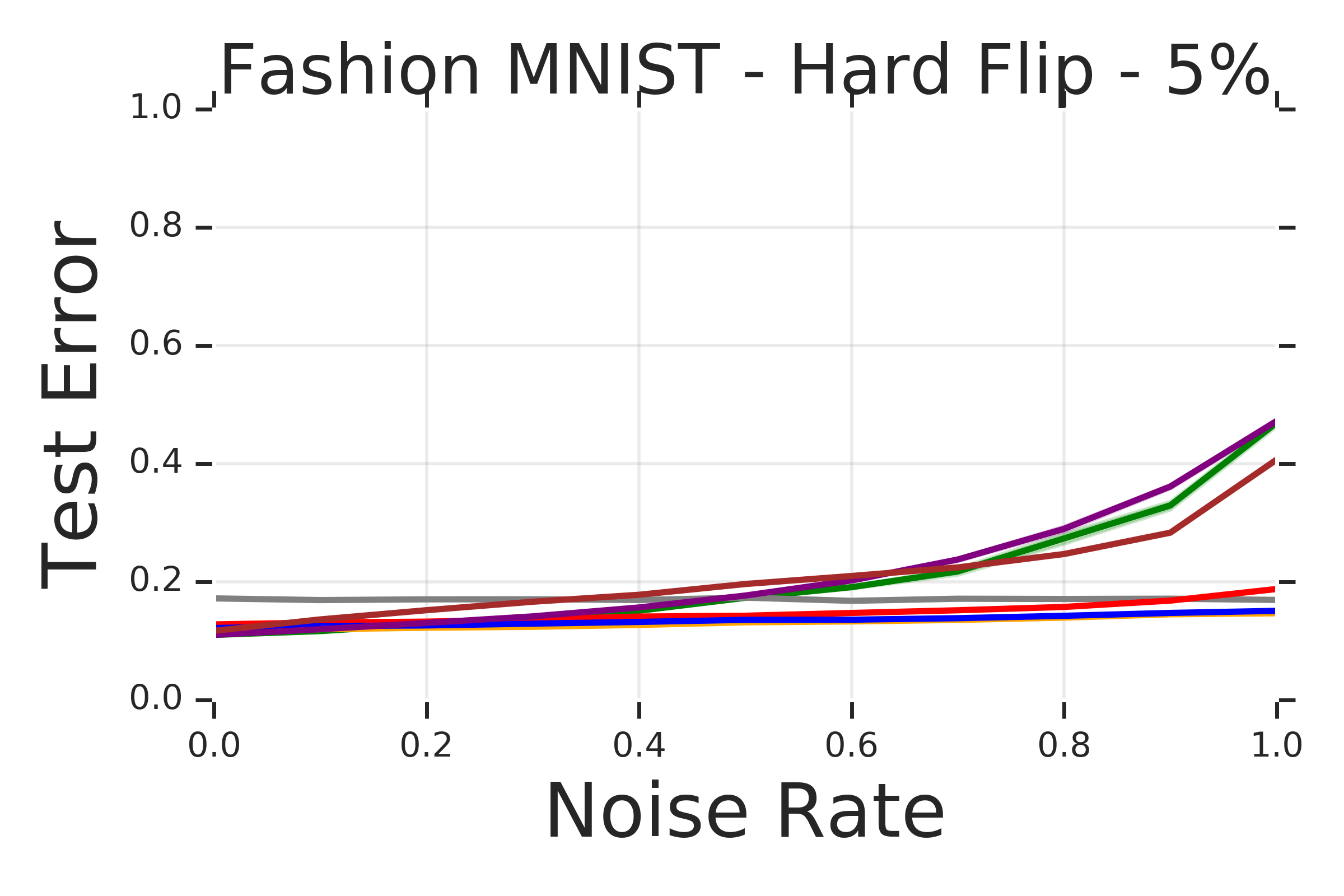}
\end{tabular}
\caption{\label{fig:uci_mnist} {\bf Top row: UCI Results, bottom row: Fashion MNIST}. Error for different amounts of noise applied to the labels of $\mathcal{D}_{\text{noisy}}$. $\mathcal{D}_{\text{clean}}$ contains $5\%$ of the data. Each column is a different corruption and each row is for a different dataset. We see that the $k$-NN method consistently chooses the best datapoints to filter leading to lower error. More results are in the Appendix.}
\end{figure*}

{\bf Auxiliary Models:} \citet{veit2017learning} propose learning a label cleaning network on trusted data by predicting the differences between clean and noisy labels. \citet{li2017learning} suggest training on a weighted average between noisy labels and distilled predictions of an auxiliary model trained on trusted data. Given a model pre-trained on noisy data, \citet{lee2019robust} boost its generalization on clean data by inducing a generative classifier on top of hidden features from the model. The method is grounded in the assumption that the hidden features follow a class-specific Gaussian distribution.

{\bf Example Weighting:} Here we make a hard decision about whether to keep a training example, but one can also adapt the weights on training examples based on the confidence in their labels.
\citet{liu2015classification} provide an importance-weighting scheme for binary classification. \citet{ren2018learning} suggest upweighting examples whose loss gradient is aligned with those of trusted examples at every step in training. \citet{jiang2017mentornet} investigate a recurrent network that learns a sample weighting scheme to give to the base model.

\subsection{$k$-nearest neighbor theory} \label{sec:relatedtheory}
The theory of $k$-nearest neighbor classification has a long history (see e.g. \citet{fix1951discriminatory,cover1968rates,stone1977consistent,devroye1994strong,chaudhuri2014rates}). Much of the prior work focuses on $k$-NN's statistical consistency properties. However, with the growing interest in adversarial examples and learning with noisy labels, there have recently been analyses of $k$-nearest neighbor methods in these settings. \citet{wang2018analyzing} analyze the robustness of $k$-NN classification and provide a robust variant of $1$-NN classification where their notion of robustness is that predictions of nearby points should be similar. \citet{gao2016consistency} provide an analysis of the $k$-NN classifier under noisy labels and like us, show that $k$-NN can attain similar rates in the noisy setting as in the noiseless setting. \citet{gao2016consistency} assume a noise model where labels are corrupted uniformly at random, while we assume an arbitrary corruption pattern and provide results based on a notion of how spread out the corrupted points are. Moreover, we provide finite-sample bounds borrowing recent advances in $k$-NN convergence theory in the noiseless setting \citep{jiang2019non} while the guarantees of \citet{gao2016consistency} are asymptotic. \citet{reeve2019fast} provide stronger guarantees on a robust modification of $k$-NN proposed by \citet{gao2016consistency}. To the best of our knowledge, we provide the first finite-sample rates of consistency for the classical $k$-NN method in the noisy setting with very little assumptions on the label noise.

\section{Deep $k$-NN Algorithm}\label{sec:algorthm}
Recall the standard $k$-nearest neighbor classifier:
\begin{definition} [$k$-NN]
Let the $k$-NN radius of $x \in \mathcal{X}$ be $r_k(x) := \inf \{ r : |B(x, r) \cap X| \ge k \}$ where $B(x, r) := \{x' \in \mathcal{X} : |x - x'| \le r \}$ and the $k$-NN set of $x \in \mathcal{X}$ be $N_k(x) := B(x, r_k(x)) \cap X$. 
Then for all $x \in \mathcal{X}$, the $k$-NN classifier function w.r.t. $X$ has discriminant function
\begin{align*} 
\eta_k(y; x) := \frac{1}{|N_k(x)|} \sum_{i=1}^n  1\left[y_i = y, \hspace{0.1cm} x_i \in N_k(x) \right],
\end{align*}
with prediction $\eta_k(x) := \argmax_y \eta_k(y; x)$.
\end{definition}

Our method is detailed in Algorithm~\ref{alg:deepknn}. It takes a dataset $\mathcal{D}_{\text{noisy}}$ of examples with potentially noisy labels, along
with a dataset $\mathcal{D}_{\text{clean}}$ consisting of clean or trusted labels. Note that we allow $\mathcal{D}_{\text{clean}}$ to be empty (i.e. in instances where no such trusted data is available). We also take as given a model architecture $\mathcal{A}$ (e.g. a 2-layer DNN with 20 hidden nodes). Algorithm~\ref{alg:deepknn} uses the following sub-routine:

\textbf{Filtering Model Train Set Selection Procedure:}
To determine whether the model used to compute the $k$-NN should only be trained on $\mathcal{D}_{\text{clean}}$ or on $\mathcal{D}_{\text{clean}} \cup \mathcal{D}_{\text{noisy}}$, we split the $\mathcal{D}_{\text{clean}}$ 70/30 into two sets: $\mathcal{D}_{\text{cleanTrain}}$ and  $\mathcal{D}_{\text{cleanVal}}$, and train two preliminary models: one on $\mathcal{D}_{\text{cleanTrain}}$ and one on $\mathcal{D}_{\text{cleanTrain}} \cup \mathcal{D}_{\text{noisy}}$. If the model trained with $\mathcal{D}_{\text{noisy}}$ performs better on the held-out $\mathcal{D}_{\text{cleanVal}}$, then output $\mathcal{D}_{\text{noisy}} \cup \mathcal{D}_{\text{clean}}$, otherwise output $\mathcal{D}_{\text{clean}}$.  

\begin{algorithm}
\caption{Deep $k$-NN Filtering}
\label{alg:deepknn}
\begin{algorithmic}[H]
   \STATE \textbf{Inputs:} $\mathcal{D}_{\text{noisy}}$, $\mathcal{D}_{\text{clean}}$ (possibly empty), $k$, $\mathcal{A}$
   \STATE Train a model $\mathcal{M}$ with architecture $\mathcal{A}$ on the output of the Filtering Model Train Set Selection Procedure.
   \STATE Train final model with architecture $\mathcal{A}$ on  $\mathcal{D}_{\text{filtered}} \cup \mathcal{D}_{\text{clean}}$.
\end{algorithmic}
\end{algorithm}

\section{Theoretical Analysis}
\label{sec:theory}
\begin{figure*}[t]
\begin{tabular}{lll}
  \includegraphics[width=0.30\textwidth]{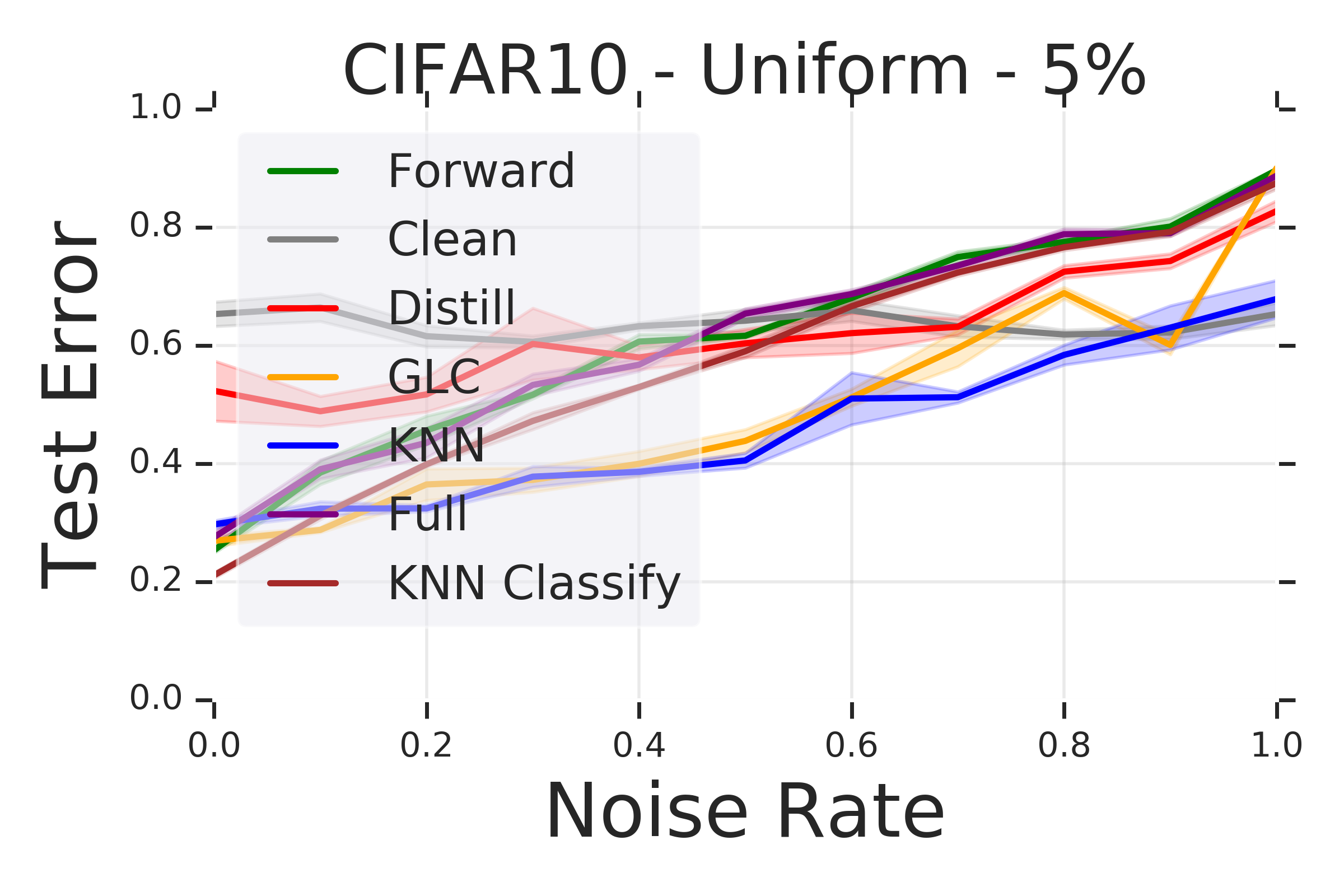} &   \includegraphics[width=0.30\textwidth]{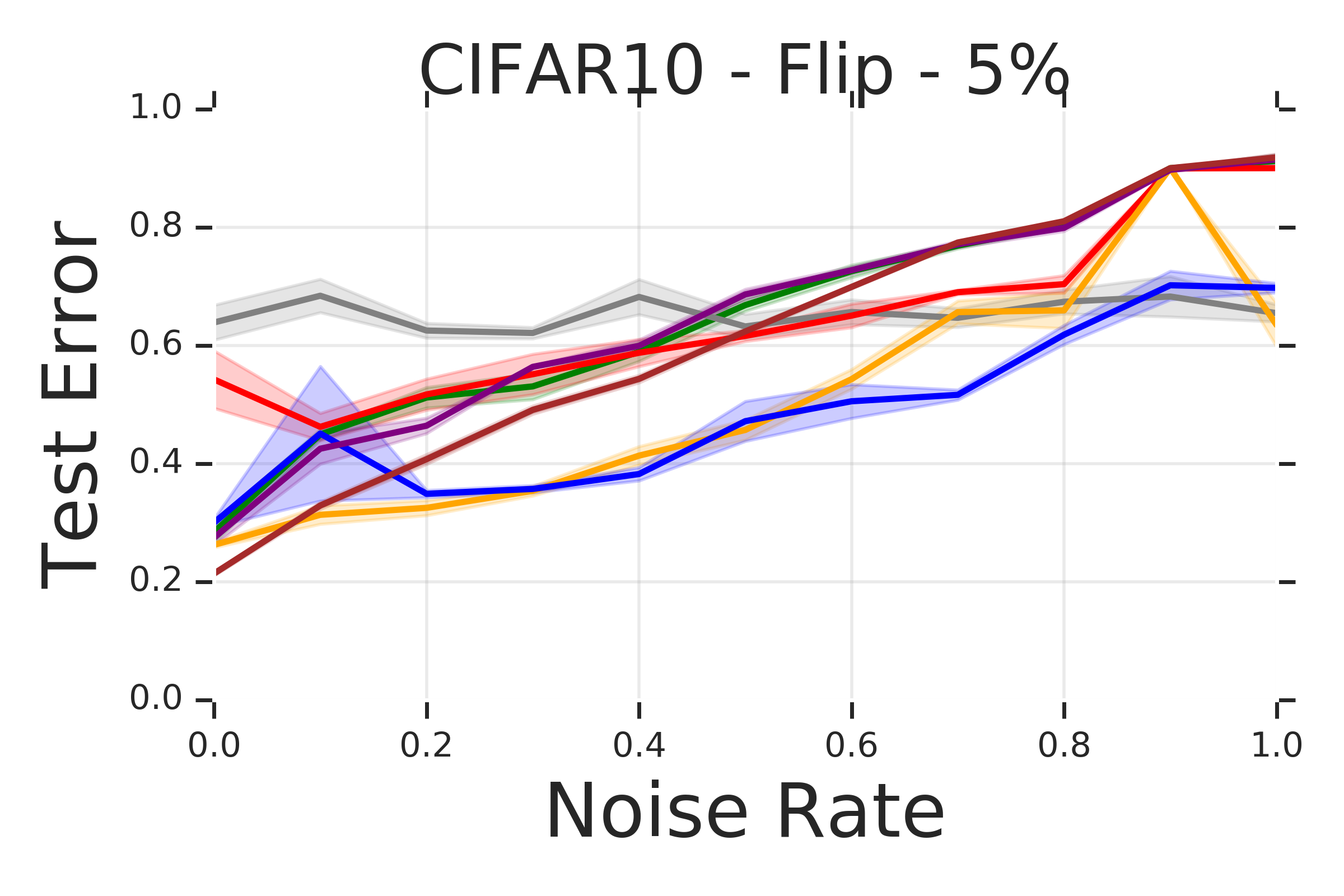} & \includegraphics[width=0.30\textwidth]{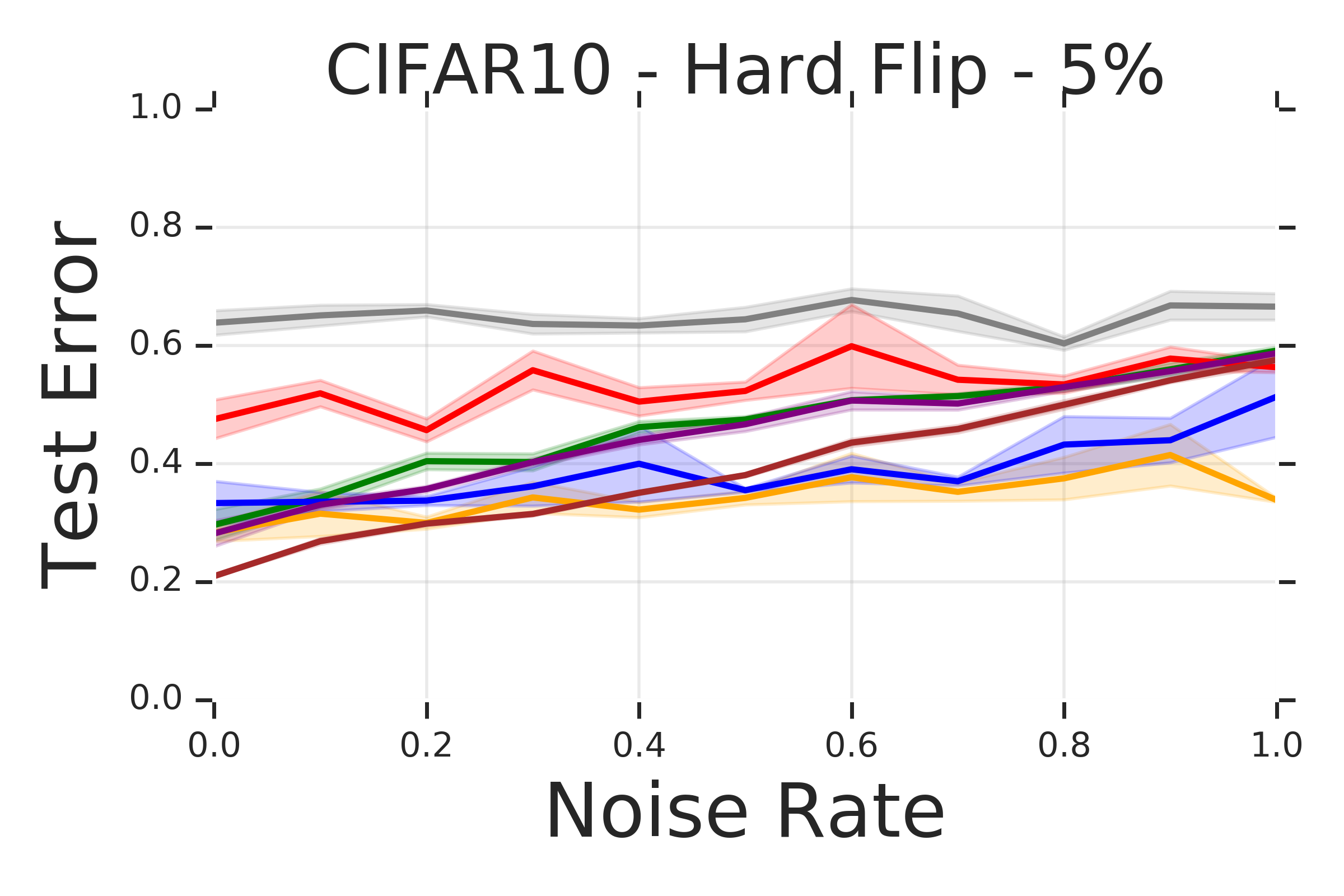} \\
  \includegraphics[width=0.30\textwidth]{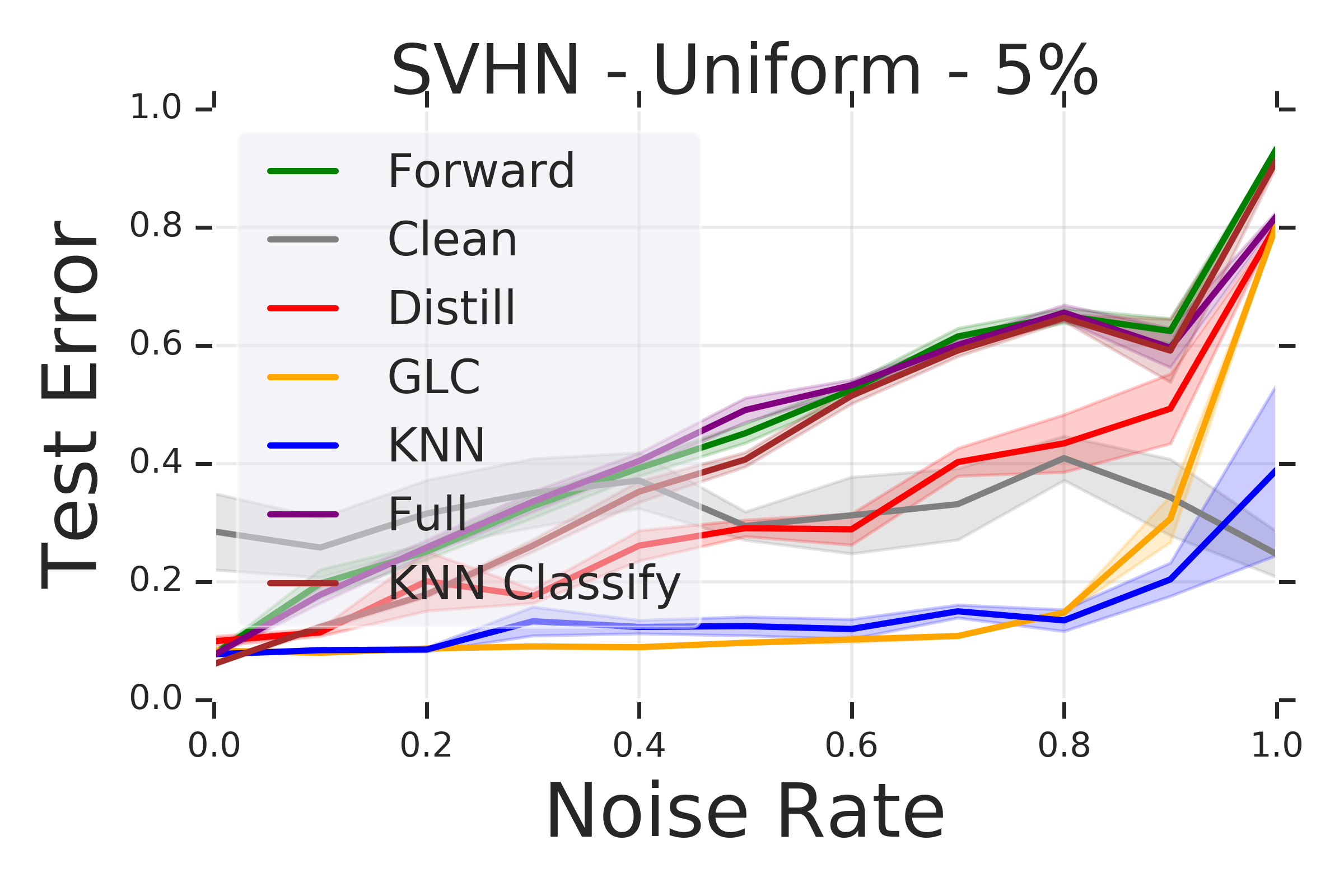} &
  \includegraphics[width=0.30\textwidth]{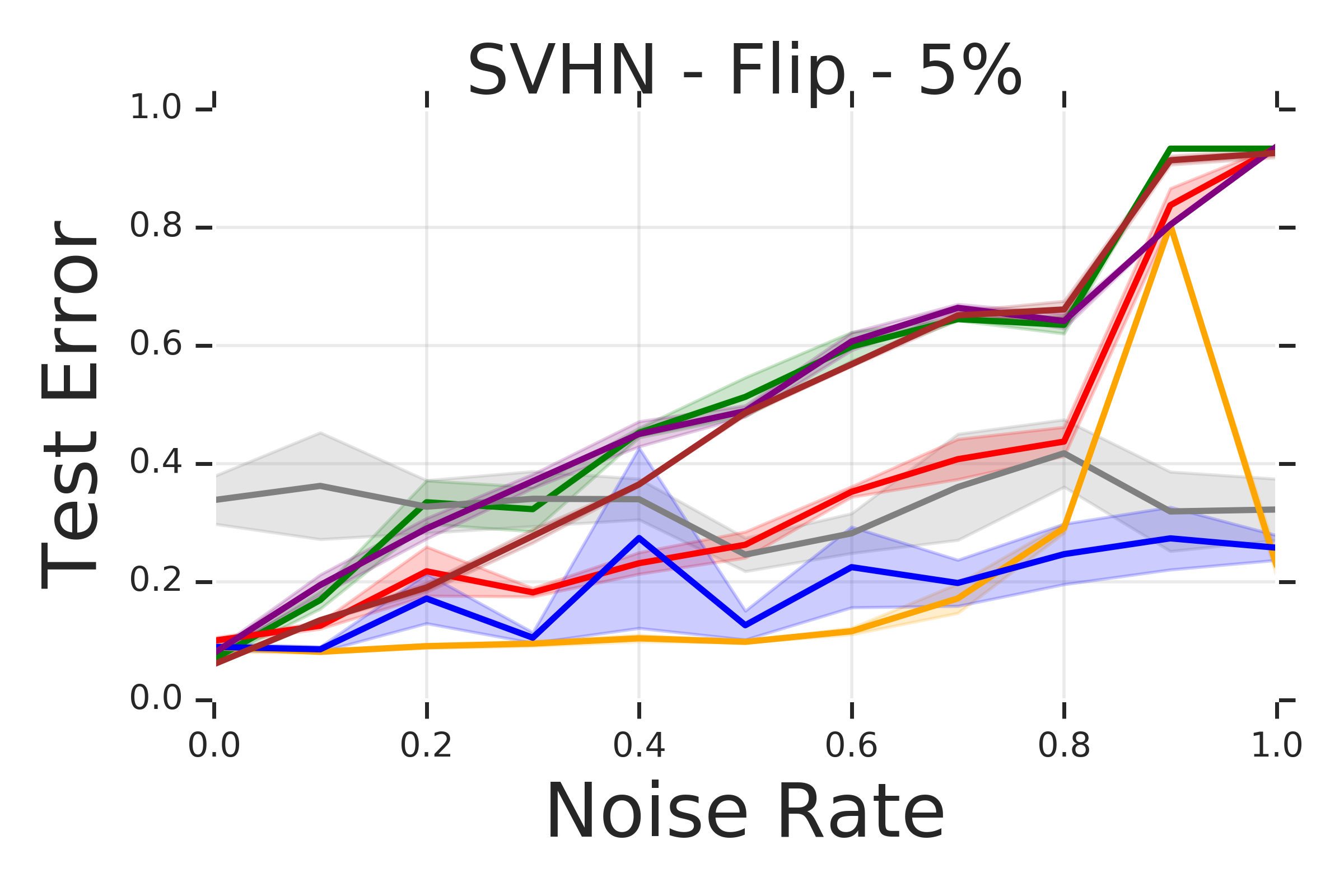} & 
  \includegraphics[width=0.30\textwidth]{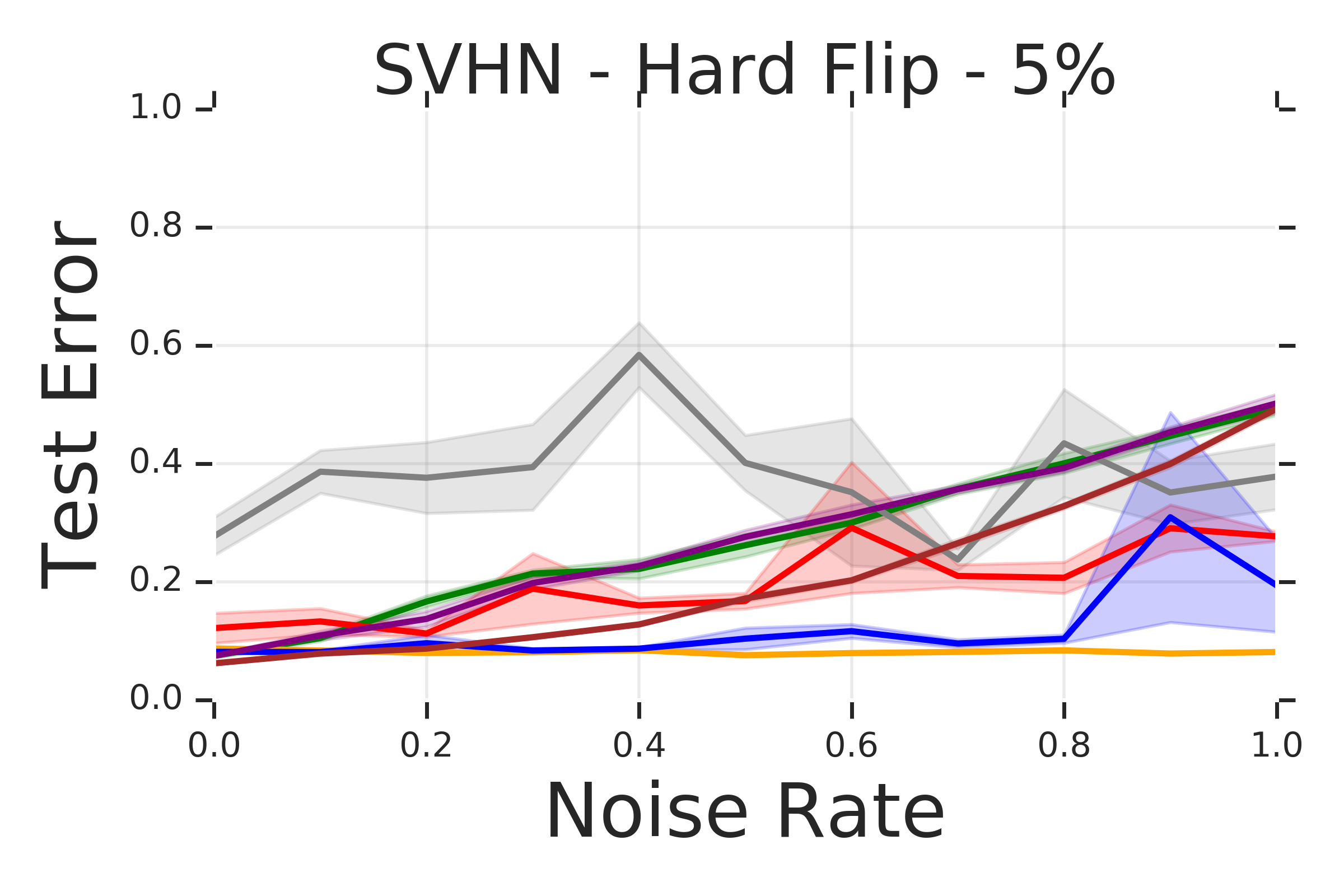} \\
\end{tabular}
\caption{\label{fig:cifar_svhn} {\bf Top row: CIFAR10, bottom row: SVHN}. Each column is a different corruption method. We see that our $k$-NN method performs competitively or outperforms on the uniform and flip noise types but performs worse for the hard flip noise type. More results are in the Appendix.}
\end{figure*}
For the theoretical analysis, we assume the binary classification problem with the features defined on compact set $\X \subseteq \R^D$. We assume that points are drawn according to distribution $\mathcal{F}$ as follows: the features come from distribution $\P_\X$ on $\mathcal{X}$ and the labels are distributed according to the measurable conditional probability function $\eta : \X \rightarrow [0, 1]$. That is, a sample $(X, Y)$ is drawn from $\mathcal{F}$ as follows: $X$ is drawn according to $\P_\X$ and $Y$ is chosen according to $\P(Y = 1|X=x) = \eta(x)$.

The goal will be to show that given corrupted examples, the $k$-NN disagreement method is still able to identify the examples whose labels do not match that of the Bayes-optimal label.

We will make a few regularity assumptions for our analysis to hold.
The first regularity assumption ensures that the support $\mathcal{X}$ does not become arbitrarily thin anywhere. This is a standard non-parametric assumption (e.g. \cite{singh2009adaptive,jiang2019non}).
\begin{assumption}[Support Regularity] \label{a1}
There exists $\omega > 0$ and $r_0 > 0$ such that $\text{Vol}(\mathcal{X} \cap B(x, r)) \ge \omega \cdot \text{Vol}(B(x, r))$ for all $x \in \mathcal{X}$ and $0 < r < r_0$, where $B(x, r) := \{x' \in \X : |x - x'| \le r\}$.
\end{assumption}

Let $p_\X$ be the density function corresponding to $\P_\X$. The next assumption ensures that with a sufficiently large sample, we will obtain a good covering of the input space.
\begin{assumption} 
[$p_\X$ bounded from below] \label{a2} $p_{X, 0} := \inf_{x \in \mathcal{X}} p_X(x) > 0$.
\end{assumption}

Finally, we make a smoothness assumption on $\eta$, as done in other analyses of $k$-NN classification (e.g. \cite{chaudhuri2014rates,reeve2019fast})
\begin{assumption}[$\eta$ H\"older continuous]\label{a3}
There exists $0 < \alpha \le 1$ and $C_\alpha > 0$ such that $|\eta(x) - \eta(x')| \le C_\alpha |x - x'|^\alpha$ for all $x, x' \in \mathcal{X}$. 
\end{assumption}

We propose a notion of how spread out a set of points is based on the minimum pairwise distance between the points. This will be a quantity in the finite-sample bounds we will present. Intuitively, the more spread out a contaminated set of points is, the less clean samples we will be needed to overcome the contamination of that set. 

\begin{definition}
[Minimum pairwise distance]
\begin{align*}
    S_2(C) := \min_{x, x' \in C, x\neq x'} |x - x'|.
\end{align*}
\end{definition}

Also define the $\Delta$-interior region of $\X$ where there is at least $\Delta$ margin in the probabilistic label:
\begin{definition}
Let $\Delta \ge 0$. Define $\X^{\Delta} := \{ x \in \X : \left|\frac{1}{2} - \eta(x)\right| \ge \Delta\}$.
\end{definition}
We now state the result, which says that with high probability {\it uniformly} on $\X^\Delta$ when $\Delta > 0$ is known, we have that the label disagrees with the $k$-NN classifier if and only if the label is not the Bayes-optimal prediction. Due to space, all of the proofs have been deferred to the Appendix.

\begin{figure*}[ht!]
\begin{tabular}{lll}
   \includegraphics[width=0.30\textwidth]{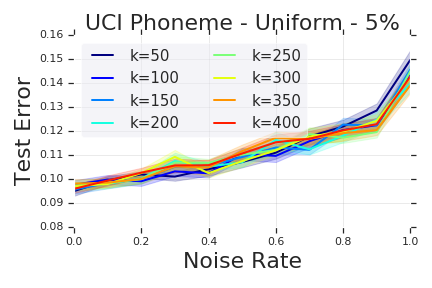} &
   \includegraphics[width=0.30\textwidth]{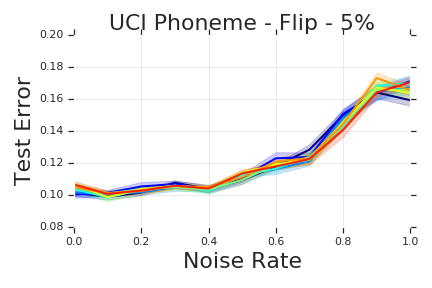} &
   \includegraphics[width=0.30\textwidth]{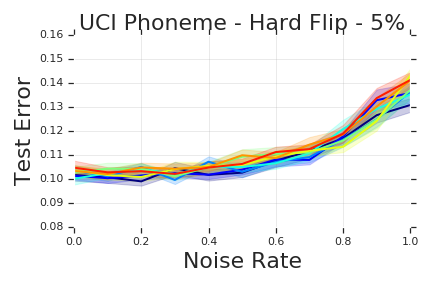} \\
\end{tabular}
\caption{\label{fig:variable_k} {\bf Performance across different values of $k$}. Here we show that on a UCI dataset, the performance of Algorithm~\ref{alg:deepknn} is stable when varying its hyperparameter $k$. Note that the $y$-axis has been zoomed in to better see the differences between the curves.}
\end{figure*}

\begin{theorem}[Fixed $\Delta$]\label{theo:fixed_delta}
Let $\Delta, \delta > 0$ and suppose Assumptions~\ref{a1},~\ref{a2}, and~\ref{a3} hold. There exists constants $K_l, K_u > 0$ depending only on $\mathcal{F}$ such that the following holds with probability at least $1 - \delta$.
Let $X_{[n]}$ be $n$ (uncorrupted) examples drawn from the $\mathcal{F}$ and $C$ be a set of points with corrupted labels and denote our sample $X := X_{[n]} \cup C$.
Suppose $k$ lies in the following range: \begin{align*}
k &\ge K_l \cdot \frac{1}{\Delta^2} \cdot \log^2(1/\delta) \cdot \log n, \\
k &\le K_u \cdot \min\{S_2(C)^D, \Delta^{D/\alpha} \} \cdot n.
\end{align*}
Then the following holds uniformly over $x \in \X^\Delta$: the $k$-NN prediction computed w.r.t. $X$ agrees with the label if and only if the label is the Bayes-optimal label $\eta^*(x) := 1[\eta(x) \ge \frac{1}{2}]$.
\end{theorem}

In the last result, we assumed that $\Delta$ was fixed. We next show how we can make a similar guarantee but show that we can take $\Delta \rightarrow 0$ as we choose $k, n \rightarrow \infty$ appropriately and provide rates of convergence.

\begin{theorem}[Rates of convergence for $\Delta$]\label{theo:rates_delta}
Let $\delta > 0$ and suppose Assumptions~\ref{a1},~\ref{a2}, and~\ref{a3} hold. There exist constants $K_l, K_u, K > 0$ depending only on $\mathcal{F}$ such that the following holds with probability at least $1 - \delta$.
Let $X_{[n]}$ be $n$ (uncorrupted) examples drawn from $\mathcal{F}$, and $C$ be a set of points with corrupted labels and denote our sample $X := X_{[n]} \cup C$.
Suppose $k$ lies in the following range: \begin{align*}
K_l  \cdot \log^2(1/\delta) \cdot n^{\frac{\alpha}{\alpha + D}} \le  k \le K_u \cdot S_2(C)^D \cdot n,
\end{align*}
then the following holds uniformly over $x \in \X^\Delta$: the $k$-NN prediction computed w.r.t. $X$ agrees with the label if and only if the label is the Bayes-optimal label $\eta^*(x) := 1[\eta(x) \ge \frac{1}{2}]$ where
\begin{align*}
    \Delta = K\cdot \left(\sqrt{\frac{\log n + \log(1/\delta)}{k}} + \left(\frac{k}{n}\right)^{\alpha/D} \right).
\end{align*}
\end{theorem}
\begin{remark}
Choosing $k = O(n^{2\alpha/(2\alpha + D)})$ in the above result gives us $\Delta = \widetilde{O}(n^{-\alpha/(2\alpha + D)})$. This rate for $\Delta$ is the minimax-optimal rate for $k$-nearest neighbor classification on $\X^\Delta$ given a sample of size $n$ \citep{chaudhuri2014rates} in the uncorrupted setting. Thus, our analysis is tight up to logarithmic factors.
\end{remark}

We next give results with an additional margin assumption, also known as Tsybakov's noise condition \citep{mammen1999smooth,tsybakov2004optimal}:
\begin{assumption}[Tsybakov Noise Condition]\label{a4} The following holds for some $C_\beta$ and $\beta$ and all $\Delta > 0$:
\begin{align*}
\P_{\X}(x \not\in \mathcal{X}^{\Delta}) \le C_\beta\cdot \Delta^{\beta}. 
\end{align*}
\end{assumption}

\begin{theorem}[Rates under Tsybakov Noise Condition]\label{theo:rates_margin}
Let $\delta > 0$ and suppose Assumptions~\ref{a1},~\ref{a2},~\ref{a3} and~\ref{a4} hold.
There exists constants $K_l, K_u, K, K' > 0$ depending only on $\mathcal{F}$ such that the following holds with probability at least $1 - \delta$.
Let $X_{[n]}$ be $n$ (uncorrupted) examples drawn from the $\mathcal{F}$ and $C$ be a set of points with corrupted labels and denote our sample $X := X_{[n]} \cup C$.
Suppose $k$ lies in the following range \begin{align*}
K_l  \cdot \log^2(1/\delta) \cdot n^{\frac{\alpha}{\alpha + D}} \le  k \le K_u \cdot S_2(C)^D \cdot n,
\end{align*}
and define $\eta_k(x) := \argmax_y \eta_k(y; x)$. Then,
\begin{align*}
\P\left(\eta_k(x) \neq \eta^*(x) \right)
    &\le  K\cdot \lambda^\beta,\\
R_X - R^* &\le   K' \cdot \lambda^{\beta + 1}, \text{where}\\
\lambda &= \left(\sqrt{\frac{\log n + \log(1/\delta)}{k}} + \left(\frac{k}{n}\right)^{\alpha/D} \right),
\end{align*}
$R_X := \E_{\mathcal{F}}[g_k(x) \neq y]$ and $R^* := \E_{\mathcal{F}}[g^*(x) \neq y]$ denote the risk of the $k$-NN method and Bayes optimal classifier, respectively.
\end{theorem}
\begin{remark}
Choosing $k = O(n^{2\alpha/(2\alpha + D)})$ in the above gives us a rate of $\widetilde{O}(n^{-\alpha(\beta + 1)/(2\alpha + D)})$ for the excess risk. This matches the lower bounds of \cite{audibert2007fast} up to logarithmic factors.
\end{remark}

\subsection{Impact of Minimum Pairwise Distance}
The minimum pairwise distance across corrupted samples, $S_2(C)$, is a key quantity in the theory presented in the previous section.
We now empirically study its significance in a simulated binary classification task in 2 dimensions. Clean samples with label $L$ are generated by sampling i.i.d from $\mathcal{N}(\mu_L, I_{2\times2})$, where $\mu_0 = (0, -2)$ and $\mu_1 = (0, 2)$. The decision boundary is the line $y = 0$. We take 100 samples uniformly spaced on a square grid centered about $(0, 0)$ and corrupt them by flipping their true label. With this construction, $S_2(C)$ is precisely the grid width, which we let vary. The training set is a union of 100 clean samples and the 100 corrupted samples. Using 1000 clean samples as a test set we study the classification performance of a majority vote $k$-NN classifier, where $k = 10$. Results are shown in Figure \ref{fig:spread}. As expected, we see that as $S_2(C)$ decreases, so does test accuracy and we need more clean training samples to compensate.

\begin{table*}[th!]
\centering
\begin{tabular}{|rccccccll|}
\hline
\multicolumn{1}{|l}{} & \multicolumn{1}{l}{} & \multicolumn{1}{l}{} & \multicolumn{1}{l}{Uniform} & \multicolumn{1}{l}{} & \multicolumn{1}{l}{} & \multicolumn{1}{l}{} &  &  \\ \hline
\multicolumn{1}{|c|}{Dataset} & \multicolumn{1}{l|}{\% Clean} & \multicolumn{1}{l|}{Forward} & \multicolumn{1}{l|}{Clean} & \multicolumn{1}{l|}{Distill} & \multicolumn{1}{l|}{GLC} & \multicolumn{1}{l|}{k-NN} & \multicolumn{1}{l|}{k-NN Classify} & Full \\ \hline
\multicolumn{1}{|r|}{} & \multicolumn{1}{c|}{5} & \multicolumn{1}{c|}{4.55} & \multicolumn{1}{c|}{3.16} & \multicolumn{1}{c|}{2.48} & \multicolumn{1}{c|}{2.33} & \multicolumn{1}{c|}{\textbf{2.05}} & \multicolumn{1}{l|}{2.19} & 2.61 \\
\multicolumn{1}{|r|}{Letters} & \multicolumn{1}{c|}{10} & \multicolumn{1}{c|}{4.28} & \multicolumn{1}{c|}{2.51} & \multicolumn{1}{c|}{2.05} & \multicolumn{1}{c|}{1.91} & \multicolumn{1}{c|}{\textbf{1.78}} & \multicolumn{1}{l|}{1.79} & 2.23 \\
\multicolumn{1}{|r|}{} & \multicolumn{1}{c|}{20} & \multicolumn{1}{c|}{3.77} & \multicolumn{1}{c|}{2.06} & \multicolumn{1}{c|}{1.76} & \multicolumn{1}{c|}{1.57} & \multicolumn{1}{c|}{1.56} & \multicolumn{1}{l|}{\textbf{1.34}} & 1.85 \\ \hline
\multicolumn{1}{|r|}{} & \multicolumn{1}{c|}{5} & \multicolumn{1}{c|}{7.89} & \multicolumn{1}{c|}{1.91} & \multicolumn{1}{c|}{1.79} & \multicolumn{1}{c|}{2.12} & \multicolumn{1}{c|}{\textbf{1.26}} & \multicolumn{1}{l|}{2.58} & 3 \\
\multicolumn{1}{|r|}{Phonemes} & \multicolumn{1}{c|}{10} & \multicolumn{1}{c|}{7.86} & \multicolumn{1}{c|}{1.54} & \multicolumn{1}{c|}{1.53} & \multicolumn{1}{c|}{1.67} & \multicolumn{1}{c|}{\textbf{1.16}} & \multicolumn{1}{l|}{2.28} & 2.85 \\
\multicolumn{1}{|r|}{} & \multicolumn{1}{c|}{20} & \multicolumn{1}{c|}{7.72} & \multicolumn{1}{c|}{1.34} & \multicolumn{1}{c|}{1.33} & \multicolumn{1}{c|}{1.35} & \multicolumn{1}{c|}{\textbf{1.13}} & \multicolumn{1}{l|}{1.76} & 2.24 \\ \hline
\multicolumn{1}{|r|}{} & \multicolumn{1}{c|}{5} & \multicolumn{1}{c|}{5.18} & \multicolumn{1}{c|}{0.56} & \multicolumn{1}{c|}{0.85} & \multicolumn{1}{c|}{0.54} & \multicolumn{1}{c|}{\textbf{0.39}} & \multicolumn{1}{l|}{0.93} & 1.86 \\
\multicolumn{1}{|r|}{Wilt} & \multicolumn{1}{c|}{10} & \multicolumn{1}{c|}{4.68} & \multicolumn{1}{c|}{0.43} & \multicolumn{1}{c|}{0.75} & \multicolumn{1}{c|}{0.45} & \multicolumn{1}{c|}{\textbf{0.32}} & \multicolumn{1}{l|}{0.77} & 1.77 \\
\multicolumn{1}{|r|}{} & \multicolumn{1}{c|}{20} & \multicolumn{1}{c|}{4.31} & \multicolumn{1}{c|}{0.36} & \multicolumn{1}{c|}{0.86} & \multicolumn{1}{c|}{0.35} & \multicolumn{1}{c|}{\textbf{0.32}} & \multicolumn{1}{l|}{0.57} & 1.5 \\ \hline
\multicolumn{1}{|r|}{} & \multicolumn{1}{c|}{5} & \multicolumn{1}{c|}{3.29} & \multicolumn{1}{c|}{4.22} & \multicolumn{1}{c|}{3.71} & \multicolumn{1}{c|}{4.2} & \multicolumn{1}{c|}{3.08} & \multicolumn{1}{l|}{2.87} & \textbf{2.71} \\
\multicolumn{1}{|r|}{Seeds} & \multicolumn{1}{c|}{10} & \multicolumn{1}{c|}{3.43} & \multicolumn{1}{c|}{3.04} & \multicolumn{1}{c|}{2.84} & \multicolumn{1}{c|}{2.99} & \multicolumn{1}{c|}{\textbf{2.14}} & \multicolumn{1}{l|}{2.74} & 2.57 \\
\multicolumn{1}{|r|}{} & \multicolumn{1}{c|}{20} & \multicolumn{1}{c|}{2.99} & \multicolumn{1}{c|}{2.69} & \multicolumn{1}{c|}{2.27} & \multicolumn{1}{c|}{2.74} & \multicolumn{1}{c|}{\textbf{1.72}} & \multicolumn{1}{l|}{2.44} & 2.2 \\ \hline
\multicolumn{1}{|r|}{} & \multicolumn{1}{c|}{5} & \multicolumn{1}{c|}{2.97} & \multicolumn{1}{c|}{3.23} & \multicolumn{1}{c|}{3.48} & \multicolumn{1}{c|}{3.97} & \multicolumn{1}{c|}{2.46} & \multicolumn{1}{l|}{\textbf{1.72}} & 2.05 \\
\multicolumn{1}{|r|}{Iris} & \multicolumn{1}{c|}{10} & \multicolumn{1}{c|}{2.89} & \multicolumn{1}{c|}{2.48} & \multicolumn{1}{c|}{2.59} & \multicolumn{1}{c|}{2.25} & \multicolumn{1}{c|}{1.32} & \multicolumn{1}{l|}{\textbf{1.26}} & 1.55 \\
\multicolumn{1}{|r|}{} & \multicolumn{1}{c|}{20} & \multicolumn{1}{c|}{2.46} & \multicolumn{1}{c|}{1.85} & \multicolumn{1}{c|}{1.97} & \multicolumn{1}{c|}{1.52} & \multicolumn{1}{c|}{\textbf{0.6}} & \multicolumn{1}{l|}{0.96} & 1.32 \\ \hline
\multicolumn{1}{|r|}{} & \multicolumn{1}{c|}{5} & \multicolumn{1}{c|}{5} & \multicolumn{1}{c|}{3.46} & \multicolumn{1}{c|}{3.26} & \multicolumn{1}{c|}{4.22} & \multicolumn{1}{c|}{3.4} & \multicolumn{1}{l|}{\textbf{3.26}} & 3.76 \\
\multicolumn{1}{|r|}{Parkinsons} & \multicolumn{1}{c|}{10} & \multicolumn{1}{c|}{5.35} & \multicolumn{1}{c|}{3.26} & \multicolumn{1}{c|}{3.22} & \multicolumn{1}{c|}{3.45} & \multicolumn{1}{c|}{\textbf{3.21}} & \multicolumn{1}{l|}{3.22} & 3.82 \\
\multicolumn{1}{|r|}{} & \multicolumn{1}{c|}{20} & \multicolumn{1}{c|}{4.88} & \multicolumn{1}{c|}{3.01} & \multicolumn{1}{c|}{3.08} & \multicolumn{1}{c|}{3.1} & \multicolumn{1}{c|}{2.98} & \multicolumn{1}{l|}{\textbf{2.97}} & 3.52 \\ \hline
\multicolumn{1}{|r|}{} & \multicolumn{1}{c|}{5} & \multicolumn{1}{c|}{2.88} & \multicolumn{1}{c|}{0.69} & \multicolumn{1}{c|}{1.03} & \multicolumn{1}{c|}{0.5} & \multicolumn{1}{c|}{\textbf{0.4}} & \multicolumn{1}{l|}{2.72} & 2.75 \\
\multicolumn{1}{|r|}{MNIST} & \multicolumn{1}{c|}{10} & \multicolumn{1}{c|}{2.57} & \multicolumn{1}{c|}{0.5} & \multicolumn{1}{c|}{0.85} & \multicolumn{1}{c|}{0.41} & \multicolumn{1}{c|}{\textbf{0.33}} & \multicolumn{1}{l|}{2.42} & 2.45 \\
\multicolumn{1}{|r|}{} & \multicolumn{1}{c|}{20} & \multicolumn{1}{c|}{2.07} & \multicolumn{1}{c|}{0.35} & \multicolumn{1}{c|}{0.69} & \multicolumn{1}{c|}{0.34} & \multicolumn{1}{c|}{\textbf{0.27}} & \multicolumn{1}{l|}{1.97} & 2.03 \\ \hline
\multicolumn{1}{|r|}{} & \multicolumn{1}{c|}{5} & \multicolumn{1}{c|}{2.76} & \multicolumn{1}{c|}{1.88} & \multicolumn{1}{c|}{1.73} & \multicolumn{1}{c|}{1.59} & \multicolumn{1}{c|}{\textbf{1.56}} & \multicolumn{1}{l|}{2.53} & 2.54 \\
\multicolumn{1}{|r|}{Fashion MNIST} & \multicolumn{1}{c|}{10} & \multicolumn{1}{c|}{2.47} & \multicolumn{1}{c|}{1.71} & \multicolumn{1}{c|}{1.6} & \multicolumn{1}{c|}{1.52} & \multicolumn{1}{c|}{\textbf{1.52}} & \multicolumn{1}{l|}{2.21} & 2.3 \\
\multicolumn{1}{|r|}{} & \multicolumn{1}{c|}{20} & \multicolumn{1}{c|}{2.07} & \multicolumn{1}{c|}{1.56} & \multicolumn{1}{c|}{1.48} & \multicolumn{1}{c|}{1.45} & \multicolumn{1}{c|}{\textbf{1.44}} & \multicolumn{1}{l|}{1.95} & 2.05 \\ \hline
\multicolumn{1}{|r|}{} & \multicolumn{1}{c|}{5} & \multicolumn{1}{c|}{6.74} & \multicolumn{1}{c|}{7} & \multicolumn{1}{c|}{6.86} & \multicolumn{1}{c|}{5.43} & \multicolumn{1}{c|}{\textbf{5.03}} & \multicolumn{1}{l|}{6.34} & 6.74 \\
\multicolumn{1}{|r|}{CIFAR10} & \multicolumn{1}{c|}{10} & \multicolumn{1}{c|}{6.58} & \multicolumn{1}{c|}{6.58} & \multicolumn{1}{c|}{6.32} & \multicolumn{1}{c|}{5.39} & \multicolumn{1}{c|}{\textbf{5.27}} & \multicolumn{1}{l|}{6.11} & 6.55 \\
\multicolumn{1}{|r|}{} & \multicolumn{1}{c|}{20} & \multicolumn{1}{c|}{6.4} & \multicolumn{1}{c|}{5.52} & \multicolumn{1}{c|}{5.66} & \multicolumn{1}{c|}{5.11} & \multicolumn{1}{c|}{\textbf{4.57}} & \multicolumn{1}{l|}{5.93} & 6.36 \\ \hline
\multicolumn{1}{|r|}{} & \multicolumn{1}{c|}{5} & \multicolumn{1}{c|}{10.8} & \multicolumn{1}{c|}{10.22} & \multicolumn{1}{c|}{9.98} & \multicolumn{1}{c|}{9.59} & \multicolumn{1}{c|}{9.57} & \multicolumn{1}{l|}{\textbf{9.17}} & 9.29 \\
\multicolumn{1}{|r|}{CIFAR100} & \multicolumn{1}{c|}{10} & \multicolumn{1}{c|}{10.79} & \multicolumn{1}{c|}{9.94} & \multicolumn{1}{c|}{9.7} & \multicolumn{1}{c|}{9.42} & \multicolumn{1}{c|}{9.63} & \multicolumn{1}{l|}{\textbf{9.09}} & 9.25 \\
\multicolumn{1}{|r|}{} & \multicolumn{1}{c|}{20} & \multicolumn{1}{c|}{10.78} & \multicolumn{1}{c|}{9.38} & \multicolumn{1}{c|}{9.15} & \multicolumn{1}{c|}{9.06} & \multicolumn{1}{c|}{9.23} & \multicolumn{1}{l|}{\textbf{8.92}} & 9.07 \\ \hline
\multicolumn{1}{|r|}{} & \multicolumn{1}{c|}{5} & \multicolumn{1}{c|}{5.04} & \multicolumn{1}{c|}{3.52} & \multicolumn{1}{c|}{3.56} & \multicolumn{1}{c|}{1.99} & \multicolumn{1}{c|}{\textbf{1.62}} & \multicolumn{1}{l|}{4.64} & 4.95 \\
\multicolumn{1}{|r|}{SVHN} & \multicolumn{1}{c|}{10} & \multicolumn{1}{c|}{4.98} & \multicolumn{1}{c|}{2.2} & \multicolumn{1}{c|}{3.2} & \multicolumn{1}{c|}{2.27} & \multicolumn{1}{c|}{\textbf{1.34}} & \multicolumn{1}{l|}{4.51} & 4.82 \\
\multicolumn{1}{|l|}{} & \multicolumn{1}{c|}{20} & \multicolumn{1}{c|}{4.32} & \multicolumn{1}{c|}{1.83} & \multicolumn{1}{c|}{2.67} & \multicolumn{1}{c|}{2.14} & \multicolumn{1}{c|}{\textbf{1.2}} & \multicolumn{1}{l|}{3.96} & 4.41 \\ \hline
\end{tabular}
\caption{\label{tab:uniform_area_under_curve} {\bf Area under the test error vs noise rate curve}. Each row corresponds to a dataset and size of clean dataset $\mathcal{D}_{\text{clean}}$ pair, where the size is a percentage of the total training set (5\%, 10\%, 20\%), for the Uniform noise type. The results for Flip and Hard Flip are in the Appendix, due to space constraints. We note that the $k$-NN method consistently outperforms the other methods for Uniform and Flip and outperforms the other methods on Hard Flip on the smaller datasets.}
\end{table*}
\section{Experiments With Clean Auxiliary Data}\label{sec:dirtyExperiments}

In this section we present experiments where a small set of relatively clean labeled data is given as side information. The methods in Section \ref{sec:cleanExperiments} do not assume such clean set is available.

\subsection{Experiment Set-up }
We ran experiments for different sizes of $\mathcal{D}_{\text{clean}}$, different noise rates, and different label corruptions, as detailed.

We randomly partition each dataset's train set into $\mathcal{D}_{\text{noisy}}$ and  $\mathcal{D}_{\text{clean}}$, and we present results for 95/5, 90/10, and 80/20 splits. 

We corrupt the labels in $\mathcal{D}_{\text{noisy}}$ for a fraction of the examples - the experiment's ``noise rate'' - using one of three schemes:

\noindent \textbf{Uniform:} The label is flipped to any one of the labels (including itself) with equal probability. 

\noindent \textbf{Flip:} The label is flipped to any \emph{other} label with equal probability.

\noindent \textbf{Hard Flip:} With probability $1/2$, we flip the label $m$ to $\pi(m)$ where $\pi$ is some predefined permutation of the labels. The motivation here is to simulate confusion between semantically similar classes, as done in \citet{hendrycks2018using}.

\subsection{Comparison Methods}
We compare against the following:

\textbf{Gold Loss Correction (GLC):} \citet{hendrycks2018using} estimates a corruption matrix by averaging the softmax outputs of the clean examples on a model trained on noisy data.

\textbf{Forward:} \citet{patrini2017making}, similar in spirit to GLC, estimates the corruption matrix by training a model on noisy data and using the softmax output for prototype examples for each class. It does not require a clean dataset like other methods.

\textbf{Distill:} \citet{li2017learning} assigns each example in the combined dataset a ``soft" label that is a convex combination of its label and its softmax output from a model trained solely on clean data.

\textbf{Clean:} Train only on $\mathcal{D}_{\text{clean}}$.

\textbf{Full:} Train on $\mathcal{D}_{\text{clean}} \cup \mathcal{D}_{\text{noisy}}$.

\textbf{$k$-NN Classify:} Train on $\mathcal{D}_{\text{clean}} \cup \mathcal{D}_{\text{noisy}}$, but at runtime classify using $k$-NN evaluated on the logit layer (rather than the model decision). This is not a competing data cleaning method, but rather it double-checks the value of the logit layer as a metric space for $k$-NN.

\subsection{Other Experiment Details}
We report test errors and show the average across multiple runs with standard error bands shaded. Errors are computed on 11 uniformly distributed noise rates between 0 and 1 inclusive. For the results shown in the main text, we have that $\mathcal{D}_{\text{clean}}$ is randomly selected and is $5\%$ of the data. In the Appendix, we show results over different sizes of $\mathcal{D}_{\text{clean}}$.
We implement all methods using Tensorflow 2.0 and Scikit-Learn. We use the Adam optimizer with default learning
rate 0.001 and a batch size of 128 across all experiments. For the UCI datasets, we set $k = 50$ and set $k = 500$ for all other datasets. We chose $k=50$ for the UCI datasets because some of the datasets were of small size. However, we found that the $k$-NN method's performance was quite stable to the choice of $k$, which we show in Section~\ref{sec:robustness_exp}. We describe the permutations used for hard flipping in the Appendix.

\subsection{UCI and MNIST Results}
We show the results for one of the UCI datasets and Fashion MNIST in Figure~\ref{fig:uci_mnist}. Due to space, results for MNIST and the remaining UCI datasets are in the Appendix. For UCI, we use a fully-connected neural network with a single hidden layer of dimension $100$ with ReLU activations and train for $100$ epochs. For both MNIST datasets, we use is a two hidden-layer fully-connected neural network where each layer has $256$ hidden units with ReLU activations. We train the model for $20$ epochs. We see that the $k$-NN approach attains models with a low error rate across noise rates and either outperforms or is competitive with the next best method, GLC.

\subsection{SVHN Results}
We show the results in Figure~\ref{fig:cifar_svhn}. We train ResNet-20 from scratch on NVIDIA P100 GPUs for 100 epochs. As in the CIFAR experiments, we see that the $k$-NN method tends to be competitive in the uniform and flip noise types but does slightly worse in the hard flip.

\subsection{CIFAR Results}
For CIFAR10/100 we use ResNet-20, which we train from scratch on single NVIDIA P100 GPUs. We train CIFAR10 for 100 epochs and CIFAR100 for 150 epochs. We show results for CIFAR10 in Figure~\ref{fig:cifar_svhn} and results for CIFAR100 in the Appendix, due to space. We see that the $k$-NN method performs competitively. It generally outperforms on the uniform and flip noise types but performs worse for the hard flip noise type. It is not too surprising that $k$-NN would be weaker in the presence of hard flip noise (i.e. where labels are mapped based on a pre-determined mapping between labels) as the noise is much more structured in that case making it more difficult to be filtered out by majority vote among the neighbors. In other words, unlike the uniform and flip noise types, we are no longer dealing with {\it white label noise} in the hard flip noise type.

\begin{figure*}[th!]
\begin{tabular}{lll}
   \includegraphics[width=0.30\textwidth]{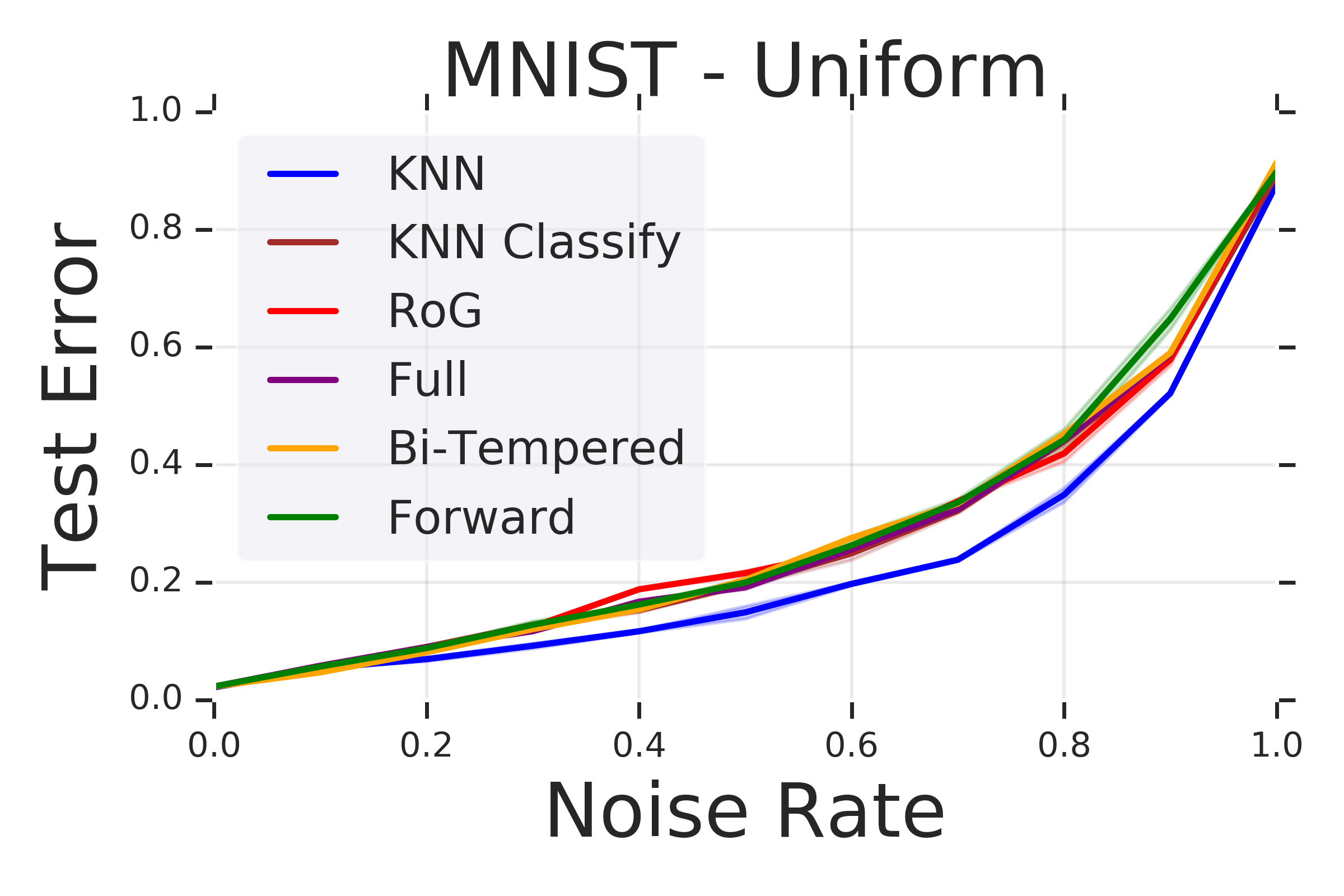} &
   \includegraphics[width=0.30\textwidth]{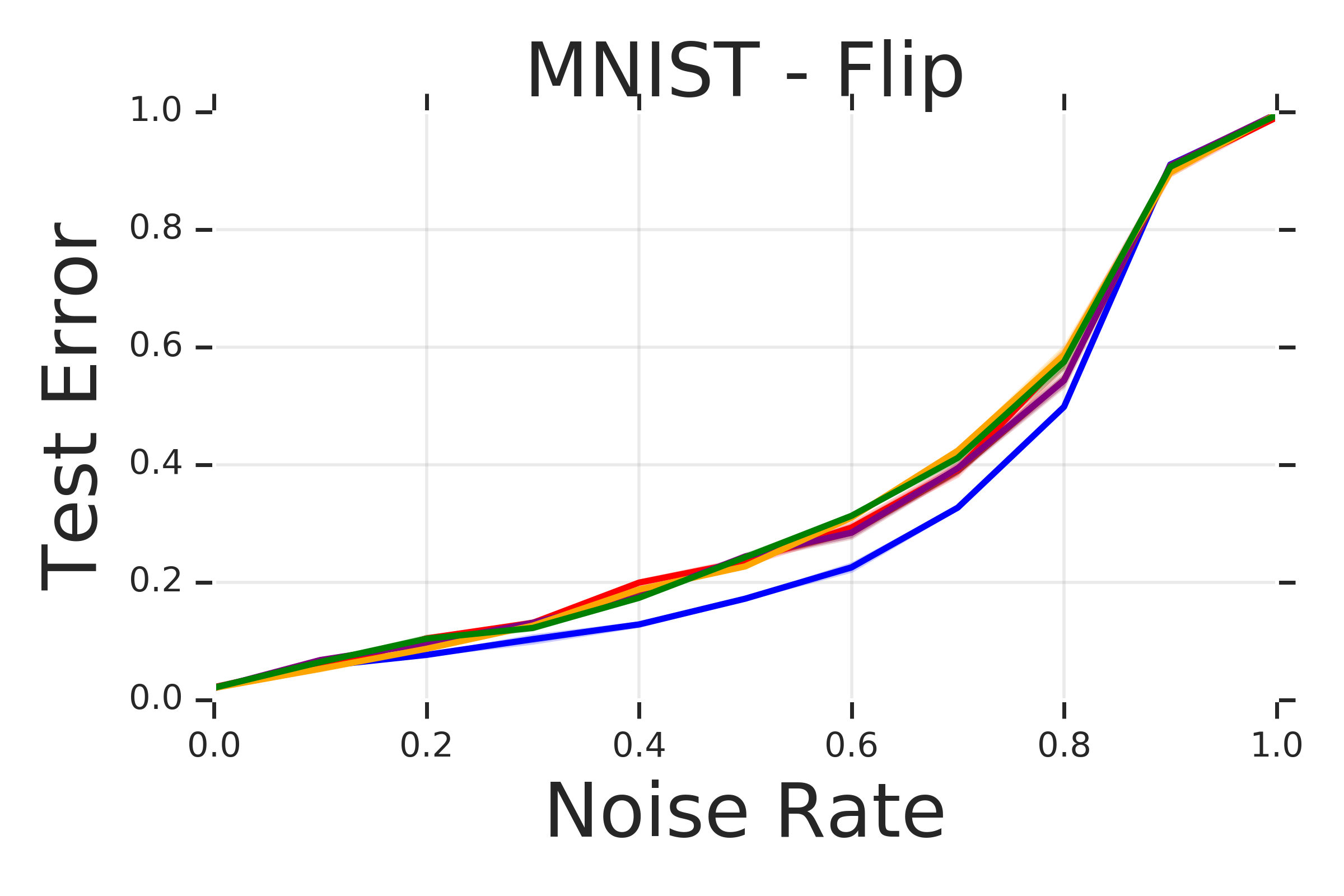} &
   \includegraphics[width=0.30\textwidth]{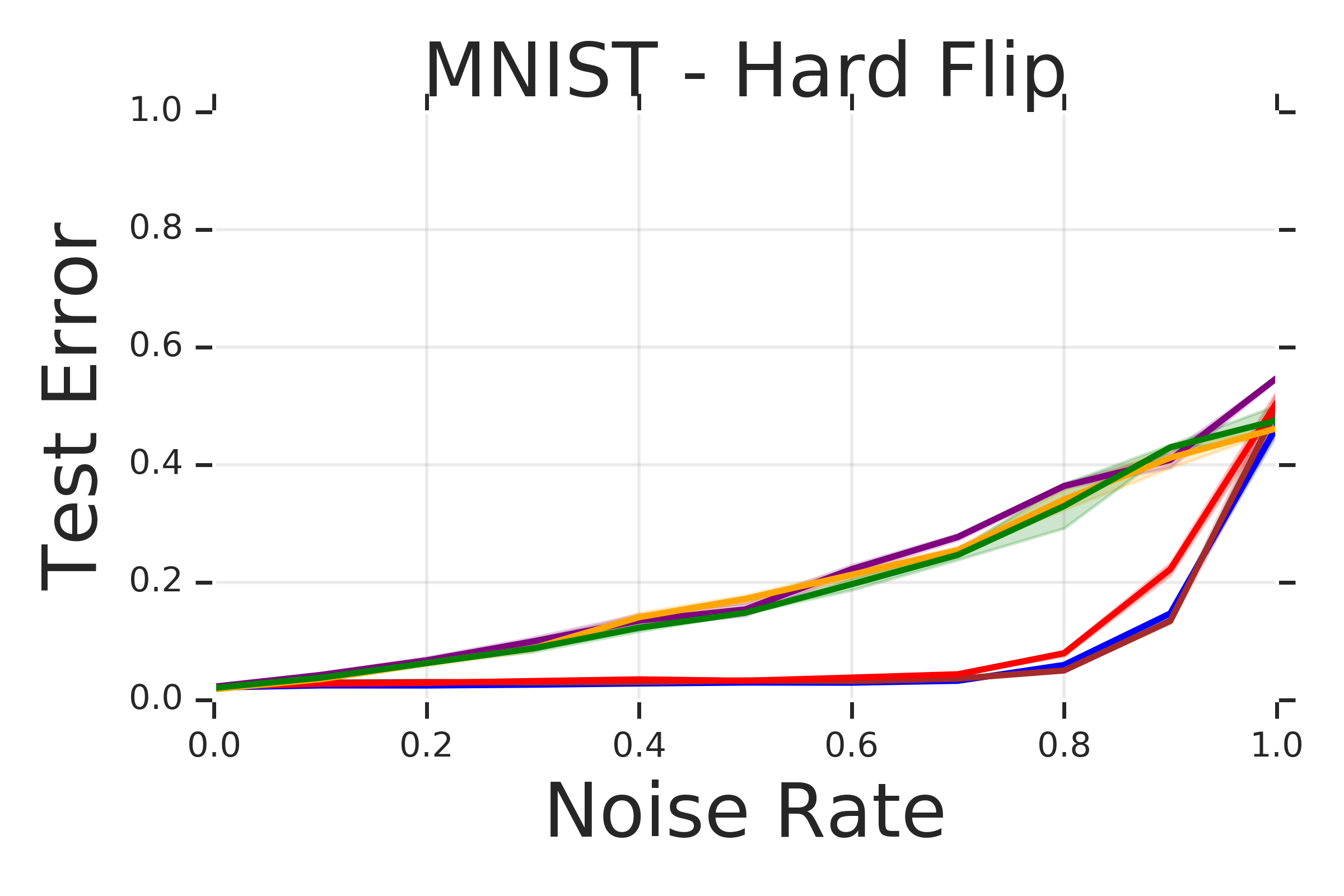} \\
   \includegraphics[width=0.30\textwidth]{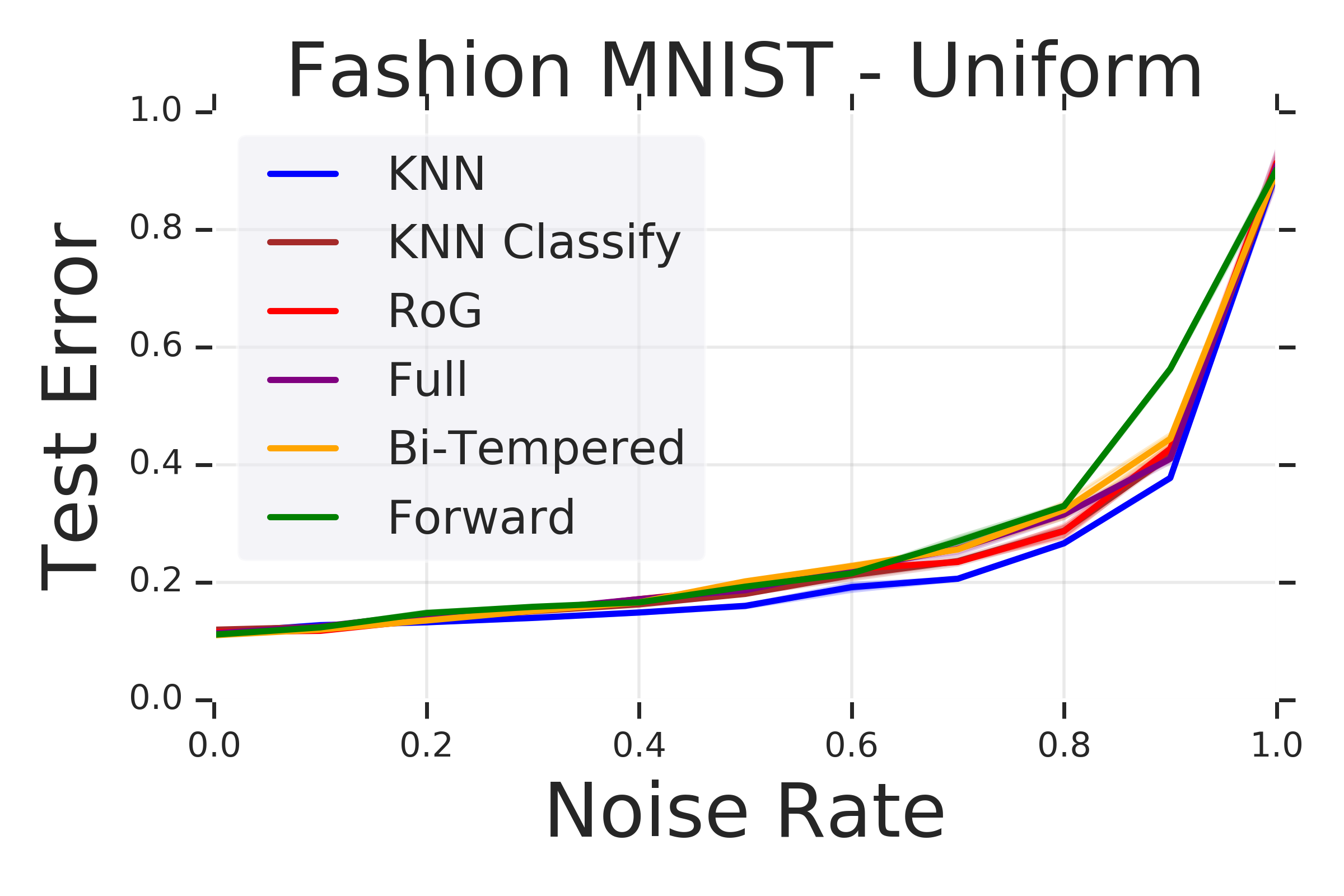} &
   \includegraphics[width=0.30\textwidth]{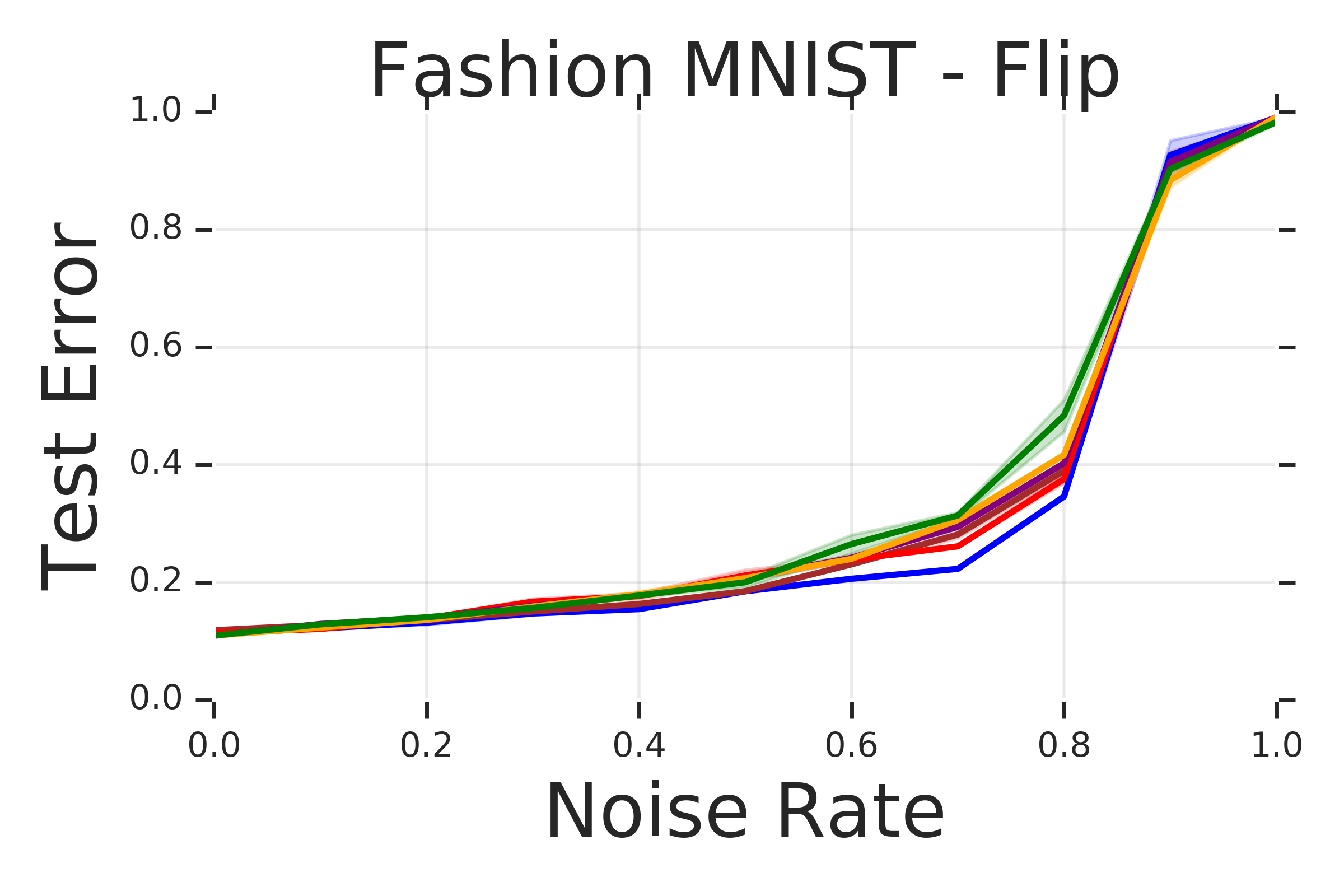} &
   \includegraphics[width=0.30\textwidth]{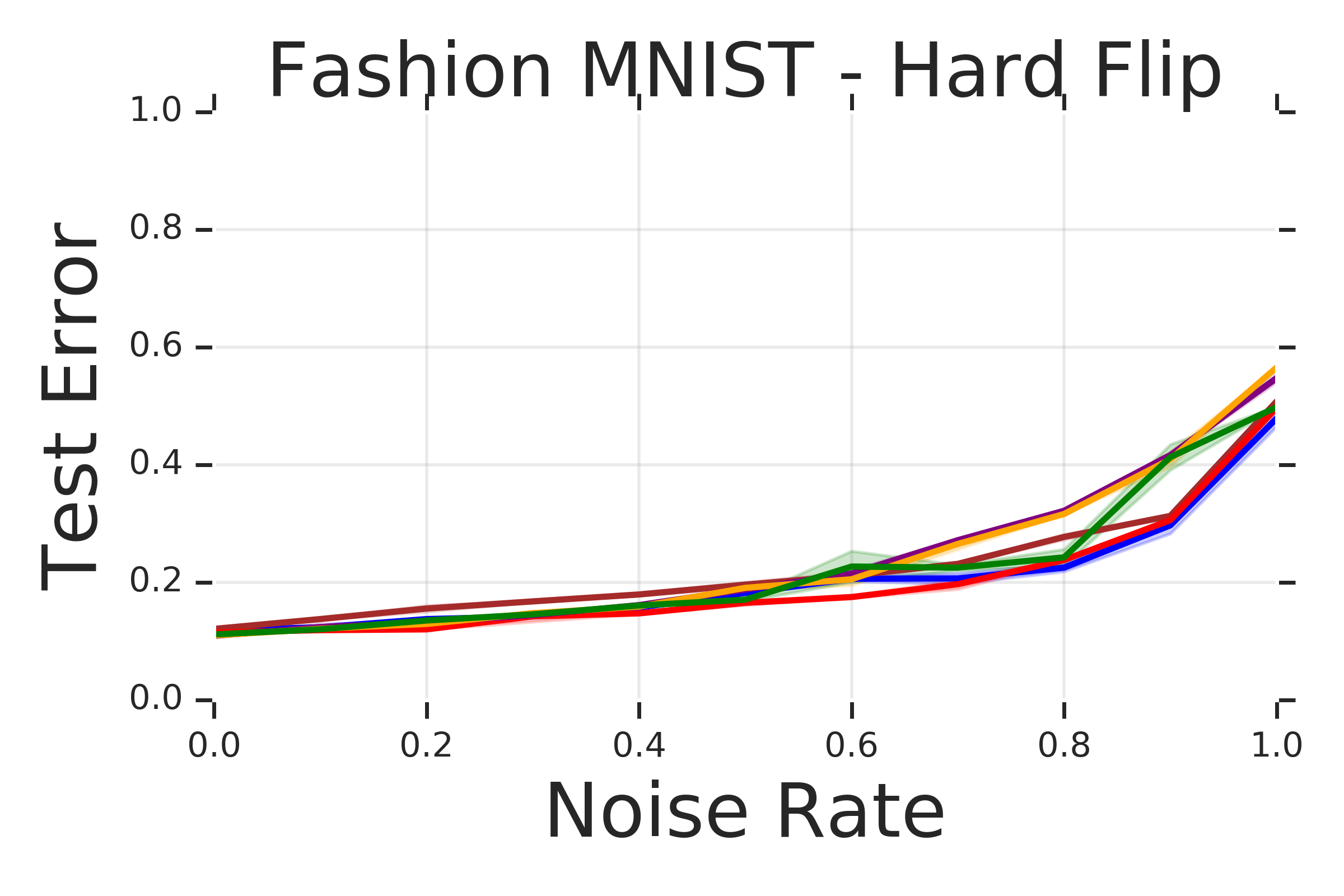} \\
\end{tabular}
\caption{\label{fig:no_clean_mnist} {\bf MNIST with $\mathcal{D}_{\text{clean}} = \emptyset$}. We find that on MNIST, KNN consistently matches or outperforms the other methods across corruption types and noise levels.}
\end{figure*}

\subsection{Robustness to $k$}\label{sec:robustness_exp}
In this section, we show that our procedure is stable in its hyperparameter $k$. The theoretical results suggest that a wide range of $k$ can give us statistical consistency guarantees, and in Figure~\ref{fig:variable_k} we show that a wide range of $k$ gives us similar results for Algorithm~\ref{alg:deepknn}. Such robustness in hyperparameter is highly desirable because optimal tuning is often not available, especially when no sufficient clean validation set is available.

\subsection{Summary of Results}
We summarize the results in Table~\ref{tab:uniform_area_under_curve}  by reporting the area under the curve for the results shown in the figures for the different datasets for the Uniform noise corruption. The best results (bolded) are generally with deep $k$-NN. Except in one case, when $k$-NN filtering does not produce the best results,  the best results are using the same $k$-NN to directly classify, rather than re-training the deep model (marked $k$-NN Classify). One reason not to use $k$-NN Classify is its slow runtime. If we insist on re-training the model and only allow the use of $k$-NN to filter, then $k$-NN loses 6 of the 33 experiments, with Full (no filtering) the 2nd best method. Analogous tables for Flip and Hard Flip are in the Appendix and show similar results. 

\section{Experiments Without Clean Auxiliary Data} \label{sec:cleanExperiments}
Next, we do not split the train set - that is,  $\mathcal{D}_{\text{clean}} = \emptyset$. The experimental setup is otherwise as-described in Section \ref{sec:dirtyExperiments}.  

\subsection{Additional Comparison Methods}
We compare against two more recently-proposed methods that do not use a $\mathcal{D}_{\text{clean}}$.

\textbf{Bi-Tempered Loss:} \citet{amid2019robust} introduces two temperatures into the loss function and trains the model on noisy data using this ``bi-tempered'' loss. We use their code from github.com/google/bi-tempered-loss and the hyperparameters suggested in their paper: $t_1 = 0.8, t_2 = 1.2$, and $5$ iterations of normalization.

\textbf{Robust Generative Classifier (RoG):} \citet{lee2019robust} induces a generative classifier using a model pre-trained on noisy data. We implemented their algorithm mimicking their hyperparameter choices - see Appendix for details.

\subsection{Results}
Results for MNIST are shown in Figure~\ref{fig:no_clean_mnist}. Due to space constraints we put results for all the other datasets presented in Section \ref{sec:dirtyExperiments} in the Appendix. The results in Figure~\ref{fig:no_clean_mnist} are mostly representative of the other experiment results. However, the results for CIFAR-100 are notable in that the logit space is 100-dimensions (since it's a 100-class problem), and we hypothesize this higher dimensional space befuddles $k$-NN a bit as it only does as well as most methods, whereas the RoG method is substantially better than most methods.

\section{Conclusions and Open Questions}
We conclude from our experiments and theory that the proposed deep $k$-NN filtering is a safe and dependable method to remove problematic training examples. While $k$-NN methods can be sensitive to the choice of $k$ when used with small datasets \citep{Garcia:2009}, we hypothesize that with today's large datasets one can blithely set $k$ to a fixed practically medium-sized value (e.g. $k=500$) as done here and expect reasonable performance. Theoretically we provided some new results for how well $k$-NN can identify clean versus corrupted labels. Open theoretical questions are whether there are alternate notions of how to characterize the difficulty of a particular configuration of corrupted examples and whether we can provide both upper and lower learning bounds under these noise conditions. 

\bibliography{paper}
\bibliographystyle{icml2020}

\end{document}